\documentclass[authoryear]{elsarticle}

\usepackage{lipsum}
\makeatletter
\def\ps@pprintTitle{%
 \let\@oddhead\@empty
 \let\@evenhead\@empty
 \def\@oddfoot{}%
 \let\@evenfoot\@oddfoot}
\makeatother

\usepackage{geometry}
\usepackage[utf8]{inputenc}
\usepackage{hyperref}
\usepackage{graphicx}
\usepackage{multirow}
\usepackage{dirtytalk}
\hypersetup{
    colorlinks=true,
    linkcolor=blue,
    filecolor=magenta,
    urlcolor=cyan,
}
\usepackage{amsbsy}
\usepackage{amsmath}
\usepackage{amsfonts}
\usepackage{xcolor}
\definecolor{highlight}{HTML}{da291c} 
\newcommand{\review}[1]{#1}

\begin{document}

\begin{frontmatter}

\title{Maximizing information from chemical engineering data sets: Applications to machine learning}

\author[icl]{Alexander Thebelt\corref{cor2}}
\author[icl]{Johannes Wiebe\corref{cor2}}
\author[icl]{Jan Kronqvist}
\author[icl]{Calvin Tsay}
\author[icl]{Ruth Misener\corref{cor1}}
\address[icl]{Department of Computing, Imperial College London, 180 Queens Gate, SW7 2AZ, UK}

\cortext[cor1]{\texttt{r.misener@imperial.ac.uk}, +44 (0) 20759 48315}
\cortext[cor2]{Co-first authors}

\begin{abstract}
  It is well-documented how artificial intelligence can have (and already is having) a big impact
    on chemical engineering.
  But classical machine learning approaches may be weak for many chemical
    engineering applications.
  This review discusses how challenging data characteristics arise in chemical engineering applications.
  We identify four characteristics of data arising in chemical engineering applications that make
    applying classical artificial intelligence approaches difficult: (1) high variance,
    low volume data, (2) low variance, high volume data, (3) noisy/corrupt/missing data,
    and (4) restricted data with physics-based limitations.
  For each of these four data characteristics, we discuss applications where these data
    characteristics arise and show how current chemical engineering research is extending the fields
    of data science and machine learning to incorporate these challenges.
  Finally, we identify several challenges for future research.
\end{abstract}


\end{frontmatter}

\section{Introduction}

Data have always played a critical role in chemical engineering applications, but recent advances
    in artificial intelligence enable new possibilities for increasing the information gained from
    chemical engineering data sets.
Previous reviews discuss technical advances relevant to chemical engineering, e.g.\ artificial
    intelligence~\citep{Venkatasubramanian2019}, machine learning~\citep{Lee2018,yan2020machine},
    optimization approaches~\citep{rios2013derivative,biegler2014multi,boukouvala2016global,ning2019optimization},
    surrogate modeling~\citep{Bhosekar2018,mcbride2019overview},
    hybrid data-driven/mechanistic modeling~\citep{von2014hybrid},
    and latent variable methods~\citep{dong2018dynamic}.
Other reviews highlight the applications and possibilities for artificial intelligence in the
    process industries~\citep{qin2019advances,shang2019data,Tsay2019,PISTIKOPOULOS2021107252}.

\begin{figure}
    \centering
    \includegraphics[width=0.6\textwidth]{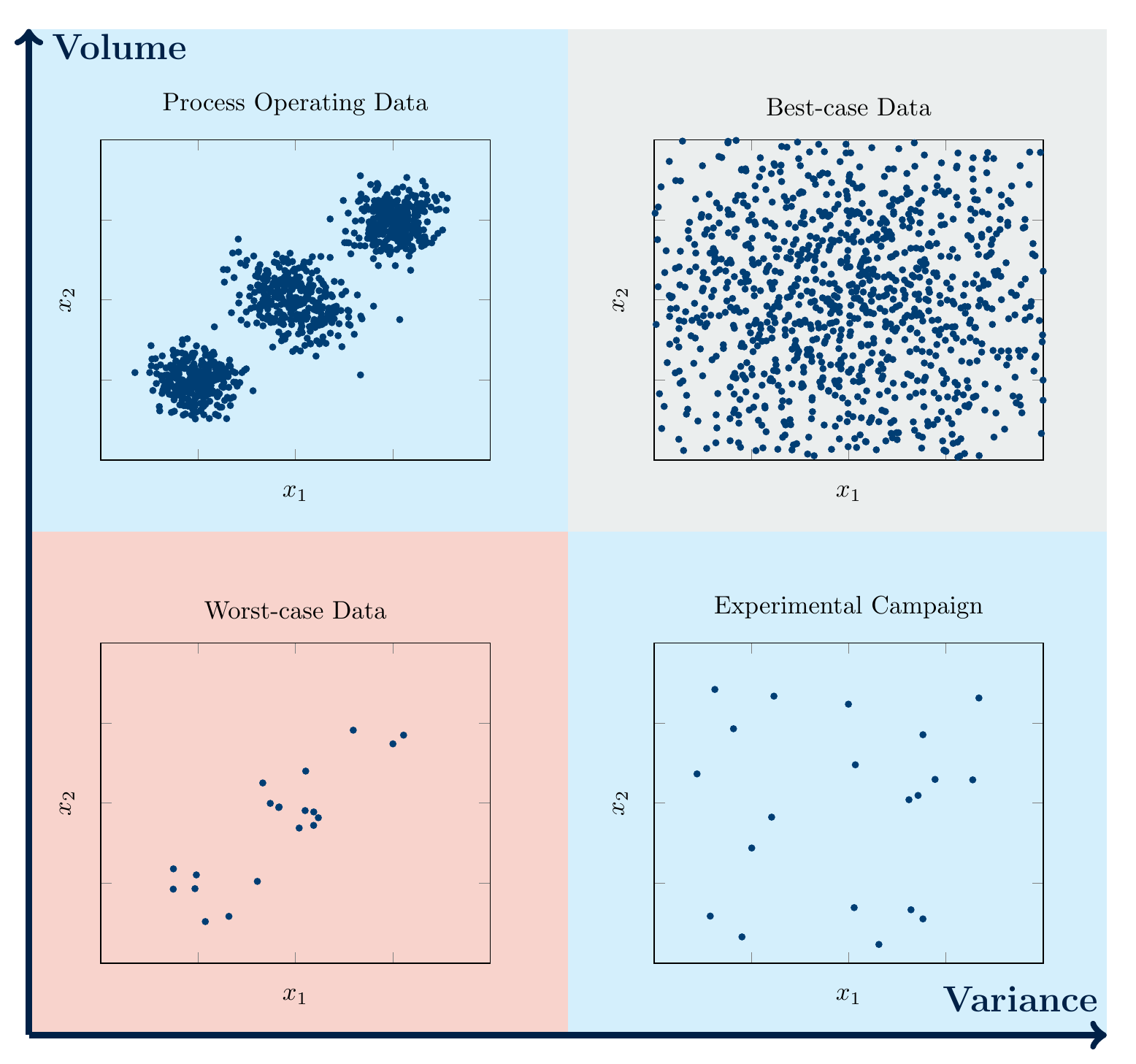}
    \caption{Variance and volume are two common descriptors characterizing data.}
    \label{fig:data_charact}
\end{figure}

This review complements previous reviews by showing how different data characteristics arise in
    chemical engineering.
Table~\ref{tab:data_characterizationl} mentions common descriptors characterizing data and gives
    examples for how these types of data arise in chemical engineering applications.
Figure~\ref{fig:data_charact} illustrates two common data concepts: variance and volume.
The upper right quadrant in Figure~\ref{fig:data_charact} (shaded grey, high variance, high volume)
    is where existing data science approaches have found most success, and this review article does
    not discuss it because classical machine learning is already relevant.
This review article discusses the upper left and lower right quadrants (shaded blue): these are
    regimes where current chemical engineering research offers transformations to increase the
    information content of relevant data sets.
We do not discuss the low variance, low volume quadrant: such data are likely not useful for machine
    learning, and engineering methods should be used to generate additional data.

\begin{table}
    \centering
    \scalebox{0.95}{
\begin{tabular}{ |r|c|c|c|c|c||c|c|c|c|c|c|c|}
\hline
\multirow{2}{7.5em}{\textbf{Data type}} & \multicolumn{3}{| c |}{\textbf{Continuous}} & \multicolumn{3}{| c |}{\textbf{Integer}} & \multicolumn{3}{| c |}{\textbf{Categorical}} &\multicolumn{3}{| c |}{\textbf{ Binary/Boolean}}\\
&\multicolumn{3}{| c |}{{\scriptsize \textit{operating conditions}}} & \multicolumn{3}{| c |}{{\scriptsize \textit{\# of processing units}}} & \multicolumn{3}{| c |}{{\scriptsize \textit{type of catalyst}}} &\multicolumn{3}{| c |}{{\scriptsize \textit{sensor status}}}\\
\hline
\multirow{2}{7.5em}{\textbf{Dimensionality}} & \multicolumn{6}{| c |}{\textbf{Time series data}} & \multicolumn{6}{| c |}{\textbf{Spatial data}}\\
& \multicolumn{6}{| c |}{{\scriptsize \textit{online measurements of physical proprieties}}} & \multicolumn{6}{| c |}{{\scriptsize \textit{spatial temperature distribution}}}\\
\hline
\multirow{2}{7.5em}{\textbf{Reliability}} & \multicolumn{4}{| c |}{\textbf{Missing data}} & \multicolumn{4}{| c |}{\textbf{Corrupted data}} & \multicolumn{4}{| c |}{\textbf{Noisy data}}\\
& \multicolumn{4}{| c |}{{\scriptsize \textit{unknown states/sensor failure}}} & \multicolumn{4}{| c |}{{\scriptsize \textit{drifting/systematic errors}}} & \multicolumn{4}{| c |}{{\scriptsize \textit{poor measurement accuracy}}}\\
\hline
\end{tabular}}
    \caption{Common descriptors that characterize data together with examples (\textit{in italics})}
    \label{tab:data_characterizationl}
\end{table}

We first discuss how challenging data characteristics arise in chemical engineering applications.
Then, we identify four characteristics of data arising in chemical engineering applications that
    make applying classical artificial intelligence approaches difficult:

\begin{itemize}
    \item \textbf{High variance, low volume data.} Illustrated in the lower right corner of
        Figure~\ref{fig:data_charact},
    \item \textbf{Low variance, high volume data.} Illustrated in the upper left corner of
        Figure~\ref{fig:data_charact},
    \item \textbf{Noisy/corrupt/missing data.} Data \emph{veracity} \review{can be construed as a
        third characteristic},
    \item \textbf{Restricted data.} Physics-based limitations give rise to restrictions in chemical
        engineering.
\end{itemize}

\noindent
\review{Note that these qualifying distinctions, e.g.\ between high variance/low volume versus
    low variance/high volume, may be problem-specific.
For example, \emph{low volume} may have a completely different meaning for safety-critical versus
    fairly innocuous processes.
Likewise, the definition of \emph{low variance} may change between online versus offline applications.
The meanings of the qualifiers \emph{high} and \emph{low} therefore only have significance when
    considering the scope of the application and the goals for using the data.}

\begin{figure}
    \centering
    \includegraphics[width=0.7\textwidth]{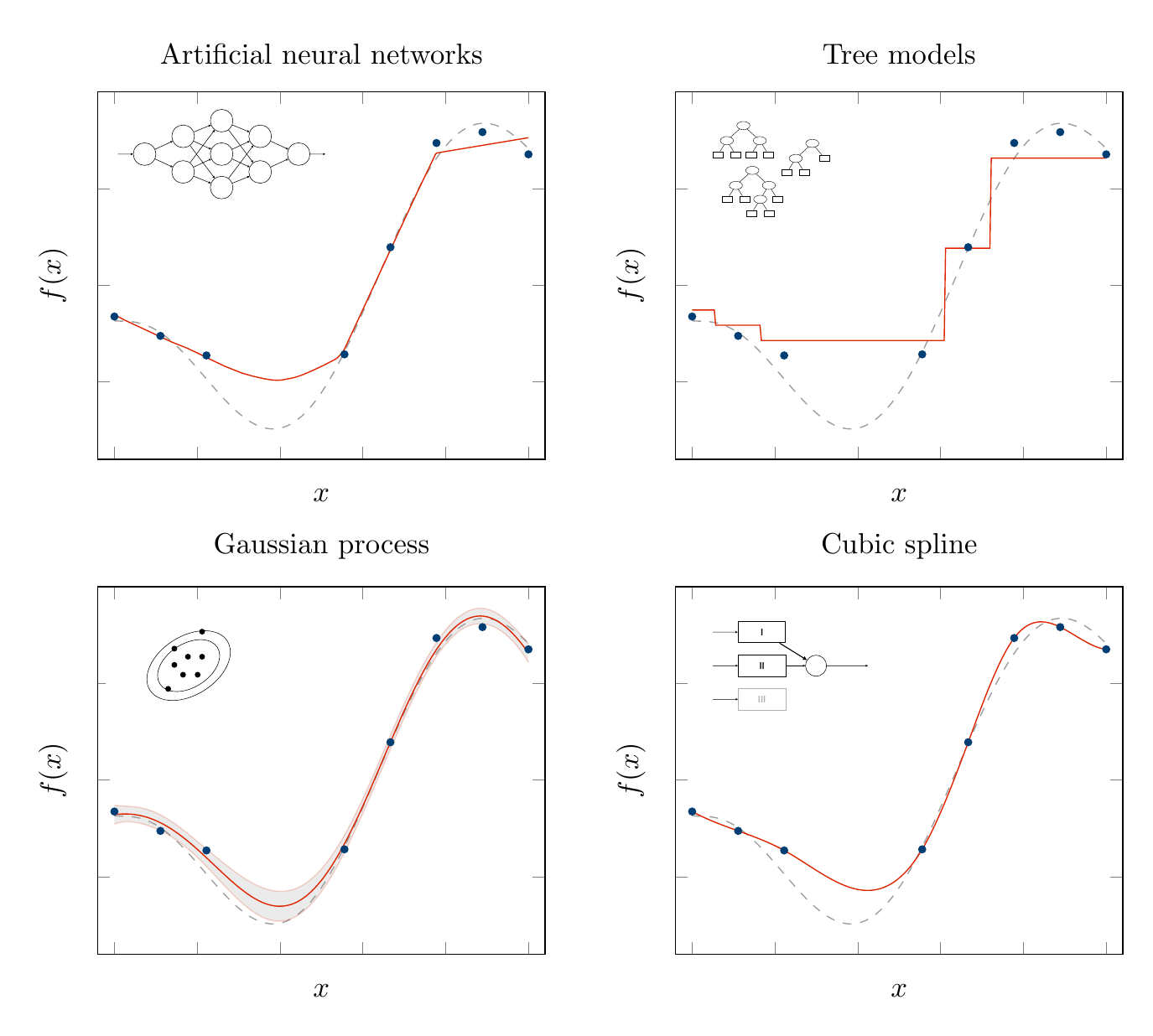}
    \caption{\review{Prediction surfaces of different types of data-driven models.}}
    \label{fig:ml_models}
\end{figure}

For each of these four data characteristics, \review{this paper discusses} applications where
    these data arise and shows how current chemical engineering research is extending the fields
    of data science and machine learning to incorporate these challenges.
We also identify challenges for future research.
The overall vision is that chemical engineers are using (1) traditional engineering approaches,
    (2) classical artificial intelligence, and (3) new research at the intersection of chemical
    engineering and artificial intelligence, to derive significant value from data that are often
    (erroneously) viewed as \textit{information-poor}.
\review{Figure~\ref{fig:ml_models} depicts several commonly used data-driven model architectures
    that we mention throughout the text.}

\section{How ``information-poor'' data arise in chemical engineering}
\label{sec:summary}

This section outlines typical characteristics of chemical engineering data sets and gives examples
    how such data sets originate in the real world.
A large number of chemical engineering applications are subject to two types of corrupted data sets,
    namely, restricted in variability, i.e.\ low variance, high volume, and restricted in volume,
    i.e.\ high variance, low volume. Sections~\ref{sec:highvar} and~\ref{sec:lowvar}, respectively,
    give a general discussion of these data.
Moreover, chemical engineering data sets may arise based on noisy, corrupted, and missing data:
    this case is presented in detail in Section~\ref{sec:noisy}. Section~\ref{sec:restricted}
    discusses various forms of inaccessible data due to known or unknown restrictions. \\

\textbf{Limited process understanding.} 
Process engineers may have limited data, e.g., from pilot or experimental campaigns, for developing
    predictive models and designing a complete process~\citep{tsay2018survey}.
As a result, the process design stage typically focuses on feasibility in meeting stringent safety,
    regulatory, and product constraints.
After finding a feasible design, process operators may be risk-averse and maintain processes close
    to a few operating points known to be feasible.
Operational changes after plant commissioning are typically minor and conducted with
    experience-guided trial and error, as changes to process set points meeting the requisite
    constraints are not easily predicted.

Similarly, there may be known physical restrictions on certain states/properties, but how these
    limitations are linked to the process parameters might not be fully understood, or they may be
    governed by complex relations.
For example, some states or properties may be described by algebraic equations, a system of partial
    differential equations (e.g., a computational fluid dynamics model), or by quantum chemistry simulations.
Process parameter restrictions may be the result of \textit{hidden constraints} on latent
    variables~\citep{martelli2014pgs,conn2009introduction}:
    for example avoiding flooding of distillation columns~\citep{gorak2014distillation} may
    introduce hidden constraints, as operators have have imperfect knowledge of the combinations of
    operating conditions and parameters that result in column flooding~\citep{piche2001flooding}.
Hidden constraints may also be fully \textit{non-quantifiable}, for which we might not be able to
    observe the outcome of certain process configurations.
For example, during data acquisition phases, non-quantifiable hidden constraints can result in
    unsuccessful experiments from which no data are obtained  or for which process simulators
    fail to converge.
Hidden constraints are especially challenging for optimization and control applications, when
    engineers cannot directly incorporate the hidden constraints into a mathematical model.
The upper left quadrant of Figure~\ref{fig:data_charact} visualizes the resulting data sets
    which are large but may be subject to low variability.
Section~\ref{sec:highvar} discusses more details of such data sets. Section~\ref{sec:restricted}
    discusses specifics of data sets arising when processes are subject to unknown limitations.
\\

\textbf{Operational limitations.}
Operators may artificially impose operational limitations when the target is process efficiency.
When a process is well understood, it can be subjected to process design optimization. 
Such design calculations traditionally assume continuous, steady-state processing, for maximizing
    the large-scale manufacturing of bulk chemical products.
However, design optimization results in sub-optimal performance when the process deviates from its
    nominal operating point, and therefore process operators seek to minimize such deviations,
    which would otherwise generate output data with more ``variance.''
Further, optimal operating points (in terms of economic performance, energy efficiency, etc.) often
    lie near one or more process constraints.
For example, it can be most efficient to make products at the requisite purity, or to operate
    equipment near safety limits.
Operation close to limits can further reduce flexibility to deviate from steady state. 

Besides process efficiency, product quality requirements can strictly limit the range of process
    parameter manipulation.
This results in large data sets with very little variation around a few operating
    conditions~\citep{qin2019advances}.
Especially in the pharmaceutical industry, there can be strict regulations regarding operation of
    production processes~\citep{troup2013process}.
We mainly consider the operational limitations to be \textit{known limitations} on the process
    parameters, giving rise to feasible domains on the process parameters.
If these limitations can be modeled mathematically as algebraic expressions, they can be
    incorporated as constraints into optimization and optimal control frameworks.
However, the exact mechanism(s) between the parameters and product properties might not always be
    fully known, and instead the limitations might be based on heuristic correlations and rules of thumb.

In addition to intentional limitations based on process efficiency and product requirements,
    there may be limitations on process operation due to equipment considerations, such as a
    maximum pressure that can be achieved by a pump/compressor, limits on heating/cooling, maximum
    achievable flow through pipes, temperature limitations due material properties, and spatial limitations.
Such limitations can also arise from safety considerations, e.g., maximum allowed temperature
    and pressure in a reactor~\citep{green2019perry}.
Limitations on process parameters and operating conditions may also be imposed by regulatory
    limits on allowed emissions, forbidding certain parameter settings.
Unreachable and unsteady states may make it practically intractable to gather data for some
    operating conditions.
For example, limitations in measurement instrumentation may restrict reliable measurements during
    relatively quick process transitions.
In experimental studies for catalyst development, certain catalyst compositions may not result in
    stable configurations usable in chemical processes.

Data sets arising from such limitations can give rise to both low variance, high volume and high
    variance, low volume data sets, see upper left and lower right quadrants in
    Figure~\ref{fig:data_charact}.
In experimental studies, such limitations may restrict search spaces, as detailed in
    Section~\ref{sec:highvar}.
On the other hand, in steady-state process settings, such limitations may limit variance of data
    sets which are still sizeable due to constant monitoring.
Section~\ref{sec:lowvar} and Section~\ref{sec:restricted} give more details.
\\

\textbf{Process control.}
In addition to guiding processes to desired states (setpoint tracking), process control systems seek
    to mitigate the effects of disturbances.
Advances in process control, e.g., improvements in controller tuning, control structures, sensors,
    continue to decrease process deviations and closed-loop response times, reducing
    disturbance-related variance in process outputs~\citep{huang1997good, simkoff2020process}.
Model predictive control (MPC), a standard technology in the chemical industry, explicitly
    accounts for process dynamics; this enables processes to transition between setpoints quickly
    and reject disturbances effectively.
Needless to say, fast disturbance rejection is desirable.
However, an unintended consequence is that variance in recorded data is further reduced, as the
    number of non-steady-state samples is likely decreased, leading to data sets similar to the
    upper left quadrant in Figure~\ref{fig:data_charact}. Section~\ref{sec:lowvar} addresses
    this in more detail. \\

\textbf{Resource constraints.}
Data gathering can be expensive: costs may be related to labor,
    material, equipment, etc.
\review{Modeling projects are often postponed or delayed to prioritize other activities,
    e.g. expansion, debottlenecking, viewed as higher impact for a production facility.}
As a result, project planners enforce tight resource budgets to limit spending. 
Moreover, certain experiments may be more expensive than others, e.g., catalyst development where
    material costs highly vary between different catalyst configurations, leading to sparse data sets.
Economic constraints can also lead to noisy, missing, and corrupted data.
Alleviating hardware problems with more accurate sensors and data transmission/storage systems
    may not be economically viable.
Often, less accurate/reliable, but cheaper sensors are used, which can lead to increased noise
    and more frequent sensor failures.
Economic constraints may also restrict the total number of sensors which can be placed in a process,
    as well as their locations, as it may be more expensive to install and maintain a sensor in
    some locations than others.
On top of economic constraints, industrial projects are subject to tight schedules that assign
    certain time frames to different project steps.
Time consuming experiments with many steps and long preparation periods may result in similarly
    sparse data sets compared to economically constrained experimental trials.

Corrupted (but useful) data sets arising from such limitations are often sparse and have high
    variability, similar to the lower right quadrant of Figure~\ref{fig:data_charact}.
Section~\ref{fig:uncertainty} discusses cases when such data sets are the result of noisy,
    missing, or corrupted measurements.
More general details on high variance, low volume data sets are given in Section~\ref{sec:highvar}. \\

\textbf{Hardware limitations.} 
Often, hardware limitations of the sensors and data transmission and storage systems lead to low
    quality data.
Measurements are inherently noisy due to interference of the environment.
Physical phenomena like sensor coking or aging can cause drift and other corruptions. 
Missing data can be the result of failed sensors or problems in acquiring, transmitting or
    storing data~\citep{Imtiaz2008}.
Measurements outside a sensor's range can also lead to missing or inaccurate values. 
Even when economic considerations are excluded and hardware is accurate, it may be impossible to
    place sensors at certain process locations because of physical restrictions.
For example, in cryogenic processes, certain sensors may not fit within the process' insulation.
Similarly, in small (e.g., intensified, microchannel) reactors, data may be limited to measurements
    from certain areas where sensors can be placed.
Hardware limitations with respect to sensors can cause both low variability in large data sets and
    limit volume of data sets with high variability, i.e., Figure~\ref{fig:data_charact} upper left
    and lower right quadrant, respectively.
Section~\ref{sec:noisy} addresses uncertain data due to hardware limitations. \\

\review{\textbf{Real-world example 1: Estimating Powder Compositions.}
As a prototypical challenge at the intersection of chemical engineering and data science, consider
    predicting a low dimensional latent structure with partial least squares for
    pharmaceuticals~\citep{doi:10.1021/acs.iecr.0c01385}.
Partial least squares outperforms principle component analysis when the output is strongly
    correlated with directions in the data that have low variance, so it may be applicable to a low
    variance regime (depending on the scope of the application).
But \cite{doi:10.1021/acs.iecr.0c01385} show an industrial example where this classical data science
    technique is insufficient: an improved approach representing the casual relationship between
    spectra and chemical composition allows more accurate predictions and shows the possibilities
    of adding chemical know-how to the data.} \\

\review{\textbf{Real-world example 2: Optimizing Catalyst Performance.}
\cite{mistry2020mixed} present another example utilizing industrial catalyst data in collaboration
    with BASF.
Modeling the effectiveness of a chemical catalyst is a tedious task that requires highly nonlinear
    models that vary between different applications.
Data-driven tree ensembles are popular for modeling such systems due to their excellent
    prediction accuracy. Moreover, such algorithms are efficient and allow for cheap model development.
\cite{mistry2020mixed} combine traditional mixed-integer techniques~\citep{mivsic2020optimization}
    to guarantee the feasibility of catalyst properties while leveraging the predictive power of
    tree ensembles trained on real industrial data.
This approach allows chemical companies to derive feasible and promising product designs while
    reducing resources dedicated towards building models.}

\section{High variance, low volume data} \label{sec:highvar}

While machine learning methods have been shown to perform well for applications where large amounts
    of data are available, it still remains challenging to get deep insights into physical processes
    given high variance, low volume data.
For this type of datasets we distinguish between two general application settings: (1) before sample
    collection, when there is a limited budget of experiments available and the goal is to maximise
    a pre-defined utility based on these trials, and (2) after sample collection, where the goal is
    to utilize a high variance, low volume dataset in the most effective way, e.g.\ for target
    prediction or process control. \\

\subsection{Relevant applications}
High variance, low volume data sets have implications for various chemical engineering applications.
The following discusses a few applications in more detail. \\

\textbf{Batch or semi-batch production.} Batch or semi-batch production processes are common when
    production demand is low and/or products are expensive.
While large chemical plants with high throughput are monitored at all times, data for batch processes
    may be limited.
Without steady-state limitations, batch procedures allow dynamic treatment, e.g.\ different
    temperature profiles, to maximize key targets such as yield and selectivity.
While this may allow better performance compared to steady-state continuous processes, it also
    makes process modeling, monitoring, and control more complicated, often leading to poorly
    performing physical models.
In the context of high variance, low volume data, applications for batch or semi-batch production
    seek to effectively use existing data rather than generating new promising data.
Ongoing research focuses on enhancing or replacing physical models with data-driven approaches.
Challenges in batch or semi-batch production processes include using high variance, low volume
    data for (1) dynamically handling controllers during experiments or optimizing control policies
    and (2) predicting final batch product quality. \\

\textbf{Material discovery and development.}
Research and development of new materials includes catalysis development, battery material production, etc.
Experimental studies optimize material configurations and properties based on single or multiple
    objectives given limited resources.
Experiments are often guided by applying physical principles, model-based design of experiments,
    and intuition derived from previous experiments.
Depending on objective, this application category is generally referred to as black-box optimization
    or design of experiments (DOE).
Challenges for these applications include: (1) deriving an initial design of experiments,
    (2) proposing promising new experiments based on previous ones, and (3) effectively exploring
    the underlying search space.
For many DOE applications, explicit constraints need to be handled to ensure feasible experimental
    settings, e.g. bounds on temperature, closure of mass-balance equations. \\

\textbf{Design of computer experiments.}
    Computational experiments can be fairly expensive and time-consuming, requiring large and powerful
    computing clusters.
Popular examples of such calculations include density functional theory, computational fluid dynamics,
    and finite element method simulations.
Such methods describe many complicated physical processes with high accuracy and may enhance, or
    even substitute for, real-world experimental studies.
Similar to real-world DOE, the main challenges include: (1) initial sets of computations to run,
    (2) proposing promising new experiments based on existing knowledge, and (3) sufficiently
    exploring the search space.
Contrary to many classic DOE settings, physical systems described by computer simulations are
    already well understood.
However, mathematical descriptions may be too complicated to be fully integrated in optimization
    problems that determine optimal experimental conditions.
In these cases, the computer simulation becomes a black-box that may be enhanced with intuition
    based on explicit constraints from the full physical model.
Optimizing these problems remains challenging. \\

\textbf{Model parameter estimation.}
Mathematical models are important for various applications in chemical engineering, e.g.\ process
    design and process control.
Many models use restrictive prior assumptions to allow a mathematical description of physical
    systems that captures major trends but is inaccurate when predicting target quantities.
Model parameter estimation based on real-world data enhances such models.
Specifically, fitting parameters in mathematical equations that capture general trends in an
    underlying system to real-world target values can improve model accuracy drastically.
In the context of high variance, low volume data, main challenges relate to (1) identifying new
    data locations to improve weaknesses in existing models, (2) estimating uncertainty of model
    parameters, and (3) deriving new governing equations from data instead of fitting parameters.\\

\subsection{How this type of data is addressed in the literature}
We categorized applications of high variance, low volume data into two categories: before and after
    sample collection.
Data-driven approaches in literature that address high variance, low volume data can be further
    categorized into (1) pre-processing of data and (2) enhancing machine learning models trained
    on low volume data.
The following presents examples of approaches that successfully handle high variance, low volume data. \\

\textbf{Feature selection and dimensionality reduction.}
Whether a dataset is low or high volume depends on the amount of data and the intrinsic problem
    dimensionality.
In contrast to dataset features, which may not accurately describe the underlying system, intrinsic
    dimensionality refers to minimum number of variables needed to represent the data.
For example, a subset of all features presented or a set of latent features (e.g., obtained via
    principal component analysis) might sufficiently represent the data.
In general, this makes machine learning models more effective as the same amount of data is
    applied to represent fewer degrees of freedom.
Therefore, dimensionality reduction algorithms identify intrinsic dimensionalities to allow
    learning in low dimensional spaces.
Such techniques are popular when deriving quantitative structure-activity relationship models using
    molecular descriptors to link structural properties to physio-chemical properties of
    interest~\citep{ponzoni2017hybridizing,eklund2014choosing}.
These type of models are used for drug discovery and optimization.
Reducing complexity without losing information is essential to obtain better fits with machine
    learning models. \cite{janet2017resolving} use feature selection techniques to improve
    accuracy of machine-learning models when predicting quantum mechanical properties for
    chemical discovery, especially in modestly sized data sets.
Other research~\citep{bartok2013representing,ghiringhelli2015big,huang2016communication} finds
    similar advantages when using feature selection techniques to model chemical properties based
    on molecular structures. \\

\textbf{Model-based and Bayesian optimization.}
Bayesian optimization (BO) is a popular approach for black-box optimization and design of experiments.
In general, BO is based on a data-driven model derived from Bayesian statistics.
For an unknown function $f: \mathbb{R}^n \rightarrow \mathbb{R}$, BO predicts the next evaluation point,
    e.g.\ the next experiment, to determine the optimal solution $x^*$ of $f$:
\begin{equation*}
    \boldsymbol{x^*} \in \underset{\boldsymbol{x} \in \mathbb{R}^n}{\text{argmin}} \; f(\boldsymbol{x})
\end{equation*}
For general black-box optimization, there is no other information available for $f$, e.g., gradients.
To derive new query points for black-box $f$, BO instead learns an approximate/surrogate model and
    optimizes an acquisition function.
Acquisition functions combine the predictive mean of the surrogate model and some variance measure
    quantifying the uncertainty of model predictions to handle the \textit{exploitation vs.\ exploration}
    trade-off.
Exploitation refines the surrogate model prediction near promising query points for black-box
    function $f$, while the exploration evaluates the underlying search-space in regions with high
    surrogate model uncertainty.
Thus, popular models for Bayesian optimization have both good prediction and uncertainty
    quantification capabilities.
Examples of such models are Gaussian processes (GPs), Bayesian neural networks, and random forests,
    \review{(see Figure~\ref{fig:ml_models} for some examples of trained models)}.
More detailed overviews of BO can be found in the
    literature~\citep{shahriari2015taking,frazier2018tutorial,lizotte2008practical}. \par
BO frameworks and data-driven model-based optimization methods have many applications in design
    of experiments.
Several model types can be used to achieve promising results. \cite{rall2019rational} show that
    artificial neural networks (ANNs) can be used to model synthetic membranes for desalination
    and ion separation processes.
The resulting models are then optimized over in a single- and multi-objective fashion to determine
    optimal fabrication conditions of the membranes used.
After carrying out the proposed experiments, the authors were able to show the reliable prediction
    ability of ANNs when modeling membrane performance. \par
\cite{bradford2018efficient} propose a multi-objective optimization method based on GPs: the
    algorithm uses spectral sampling to approximate drawn function samples of the GP posterior distribution.
A genetic algorithm optimizes the samples to find promising new evaluation points of the black-box function.
The \cite{bradford2018efficient} approach has been applied to continuous
    flow chemistry~\citep{schweidtmann2018machine} and pharmaceutical processes~\citep{clayton2020automated}.
Other GP-based approaches were applied for tissue engineering~\citep{olofsson2018bayesian},
    solar cell material optimization~\citep{herbol2018efficient}, and optimization of sustainable algal
    production~\citep{bradford2018dynamic}.
Other data-driven model-based design of experiment approaches use
    tree ensembles~\citep{mistry2020mixed,thebelt2020entmoot,thebelt2022multi} and algebraic
    basis functions~\citep{wilson2017alamo} to predict new promising points for evaluation. \\

\textbf{Hybrid modeling.}
Small datasets with high variance are useful to recognize trends in the underlying feature space
    but may be difficult to use for interpolation between data points.
Hybrid data-driven/mechanistic methods try to fill these dataset gaps with domain knowledge such
    as mathematical equations.
Here, machine learning models using small datasets can enhance existing physics-based models.
\cite{rall2020simultaneous} use data-driven models to enable both membrane synthesis and membrane
    process design in a hybrid modeling approach.
While the process design is modelled using common mechanistic models, membrane properties are
    estimated using ANNs.
They demonstrate the proposed hybrid modeling strategy can lead to better overall performance
    compared to conventional approaches that only design processes based on a collection
    of known membranes. \par
\cite{henao2011surrogate} use ANNs to replace complicated models for operation units in process
    synthesis, while keeping deterministic models for simple units.
The authors investigate applications related to design of continuous stirred tank reactors,
    solvent regeneration units, and synthesis of a reaction separation, finding that that ANNs can
    lead to compact and accurate representations of superstructure optimization models.
A key advantage is replacing various nonlinearities stemming from process models with a single type
    of nonlinearity from the activation function of the ANNs used, which enables universal treatment
    when optimizing over such models.
\cite{schweidtmann2019deterministic2} propose a framework that allows deterministic global
    optimization of ANN embedded structures.
This framework has been applied to hybrid modeling applications including optimization of
    organic Rankine cycles~\citep{huster2019working,schweidtmann2019deterministic}. \\

\textbf{Low volume data models.}
While some methods handle small data sets by increasing their volume via additional experiments,
    other approaches enhance limited data by using deterministic models.
If these options are unavailable, specifically tailored models may perform well in small data settings.
GPs are popular models for this category as they smoothly fit existing data points and give valuable
    uncertainty quantification in unexplored regions~\citep{rasmussen2003gaussian}.
While this does not solve the problem of inaccurate interpolation between distant data points,
    it does indicate where low-accuracy of the model is expected.
While GPs have a built-in uncertainty measure, there exist approaches that aim to estimate
    uncertainty for other commonly-used data-driven models.
\cite{springenberg2016bayesian} use Hamiltonian Monte Carlo methods, specifically based
    on \citet{chen2014stochastic}, to derive reliable uncertainty estimates for ANNs in low volume
    data applications.
\cite{garnelo2018conditional} propose \textit{Neural Processes} that combine the advantages of both
    GPs and ANNs and give explicit uncertainty estimates.

\section{Low variance, high volume data}
\label{sec:lowvar}

Another challenge faced in applying data-driven solutions to chemical processes is that the vast
    majority of operating data recorded by large-scale processes correspond to normal operation at a
    (few) routine point(s)~\citep{qin2019advances}.
Several reasons for this are summarized in Section~\ref{sec:summary}.
Nevertheless, the \textit{volume} of such data can be significant: a chemical process can have many
    sensors that record measurements at frequencies in the order of minutes, and stored records can
    go back in time a decade or more.

\subsection{Relevant Applications}
Machine learning methods applied to the low variance, high volume data from large-scale processes
    must be carefully selected/adapted to meet this unique challenge.
\\

\textbf{Outlier/fault detection.}
Machine learning can be deployed online to detect (and/or classify) faults or outliers in the
    behavior of a process.
Specifically, when new data are continually recorded, fault detection seeks to answer: are the
    data (statistically) different from normal operation?
In a data-driven methodology, ``normal operation'' is quantified using a recorded dataset. 
Due to its importance and direct applicability to process operations, data-driven fault detection
    has been widely studied~\citep{venkatasubramanian2003review, qin2012survey, ge2013review, jiang2019review}.
The main challenges related to low variance, high volume data are: (1) building classification
    models from unbalanced datasets (2) determining statistical limits for ``normal operation,'' and
    (3) attributing outliers/faulty measurements to a root cause.
\\

\textbf{Process drift.}
A related challenge to fault detection is identifying ``drift'' in process
    behavior~\citep{lee2011relay,montgomery1994integrating}.
For instance, closed-loop process behavior can change slowly over time due to degradation of
    control system performance (e.g., plant-model mismatch), equipment deterioration
    (e.g., heat exchanger fouling, build-up of trace components), etc.
Unlike in many fault detection applications, drifting process behavior often results in measured
    data that are not statistical outliers, but rather correspond to gradual shifts.
For equipment degradation in particular, many research efforts focus on
    \textit{condition-based monitoring}, where additional data streams specific to equipment, e.g.,
    machine vibration, audio/video data, are used to model their condition(s).
The challenges are: (1) leveraging and combining information from diverse data streams,
    (2) quantifying a slow change in the underlying process dynamics, and (3) isolating individual
    effects from other phenomena/noise in recorded data.
\\

\textbf{Flexible operations.}
Recent trends towards flexible operation of chemical processes are based on motivations to deviate
    from the accepted paradigm of steady-state operation~\citep{riese2020challenges, pattison2016optimal}.
For example, large fluctuations in electricity prices may motivate over-production when prices are low,
    and vice versa (assuming products can be stored).
Creating these flexible production schedules requires a good understanding of feasible process
    operating points, and perhaps also of feasible transitions.
Therefore, machine learning models can help production schedulers understand the feasible operating
    regimes of a process by examining data from its past operation.
When process operational data include multiple operating points, the range of feasible operating
    points and transitions can potentially be identified from historical data.
The main challenges identified here are: (1) identifying recorded data that correspond to routine
    or desirable operations and (2) creating data-driven formulations that both accurately describe
    process operations and are amenable to scheduling formulations.
\\

\textbf{Unconventional control strategies.}
As noted above, improvements to control technologies have greatly reduced variance in process data. 
However, it can at times be favorable to sacrifice some degree of control performance in order
    to gain data-driven process knowledge~\citep{mesbah2018stochastic, hewing2020learning}.
For instance, additional (multi-objective) terms can be included in a model-predictive-control
    objective function to encourage excitation of a process when model improvements are desired.
Alternatively, a reinforcement-learning-type approach can be taken, allowing the controller to
    learn an optimal policy over time rather than requiring an accurate open-loop model a priori.
Strategies such as these balance control performance and exploration of the input-output space,
    allowing process models and/or their control systems to be improved as data are collected.
The main challenges identified here are: (1a) training/improving dynamic models from
    quasi-stationary data, (1b) training/improving open-loop dynamic models with closed-loop data,
    i.e., \textit{closed-loop identification}, and (2) balancing risk-averse operation with
    exploration for model/controller improvements.
\\

\subsection{How are these challenges addressed?}
A central theme in dealing with these data is making the most of data predominantly corresponding
    to ``normal'' operation.
Intuitively, while large data volume facilitates estimation of noise/variance, low ``variance'',
    i.e., changes in operation, complicates the understanding of underlying process dynamics.
Therefore, many chemical engineering applications dealing with such data exploit subject-matter
    expertise to assist in modeling underlying behavior.
When such information is not readily available, some purely data-driven approaches have still
    found success.
\\

\textbf{Data reconciliation and moving horizon estimation.}
When a mathematical model is available in addition to process data, model predictions can differ
    significantly from measured values, owing to model assumptions, measurement error, etc.
To this end, \textit{data reconciliation} adjusts measured data, e.g., by solving a maximum
    likelihood estimation (MLE problem subject to known model equations, often comprising
    conservation laws and/or variable constraints~\citep{do2018collection}.
When a more sophisticated process model is available, its parameters can be simultaneously estimated
    via MLE, resulting in \textit{error-in-variables} problems, e.g.,
    \cite{esposito1998global, gau2002deterministic}.
Solution of this (global) optimization problem is known to be challenging: to maintain
    computational tractability, the number of considered data samples can be fixed, known as
    \textit{moving horizon estimation}~\citep{johansen2011introduction}.
Several research efforts~\citep{zavala2008fast, alessandri2011moving, hashemian2015fast} have
    focused on expediting computational solution of the MHE problem, for deployment in online applications.
In summary, data reconciliation and MHE techniques use data to continually improve process
    knowledge via a hypothesized mathematical model.
In turn, this results in improved models for process control, and MPC systems can exploit the
    reconciled measurements and parameter estimates from MHE~\citep{huang2010fast, voelker2013simultaneous}.
Additionally, fault detection can also be performed by tracking parameter estimates from
    MHE~\citep{bemporad1999moving, spivey2010constrained}.
\\ 

\textbf{Filtering and state estimation.} 
In a similar vein to the above, observed data can be used to estimate the values of the state, or
    hidden/unmeasured, variables of a process.
Specifically, the \textit{state estimation} problem determines the values of the process states,
    given a model structure and sequence of measured data.
While state estimation can also be formulated as an MHE (optimization) problem, several filtering
    methods are commonly used to update state estimates using only the most recent measurement,
    summarizing previous data using, e.g., state and covariance matrix estimates.
The popular Kalman filter provides optimal estimates for the case of a linear, unconstrained
    system subject to Gaussian noise (typically also estimated from data).
Extensions such as the extended and unscented Kalman filters enable state estimation in nonlinear
    systems by linearizing the system around its current state or additional sampled points, respectively.
Rather than assuming Gaussian noise, a sampling/Monte Carlo approach can be taken---a technique
    known as particle filtering~\citep{zhao2014constrained}.
\cite{daum2005nonlinear} and \cite{rawlings2006particle} provide more comprehensive overviews of
    (nonlinear) state estimation.
Importantly, while they are generally simpler than MHE problems, filtering-based schemes address
    similar applications, such as fault detection~\citep{bhagwat2003multi}.
Finally, Kalman-filtering-based approaches can also update estimates of model parameters by treating
    the parameters themselves equivalently to hidden process states~\citep{ljung1990adaptation, guo1990estimating}.
\\

\textbf{Scale-bridging models.}
While low variance data are often insufficient to construct a detailed process model, useful coarse
    approximations may still be derived.
For instance, several works ``bridge'' multiple time scales by identifying feasible operating regimes
    from historical data, in order to embed lower-level process knowledge in decisions at a higher
    level~\citep{Tsay2019}.
This can involve identifying samples corresponding to feasible steady-state operating
    points~\citep{xenos2016, xenos2016b} and/or modeling regions of such feasible operation using
    convex-region surrogate models~\citep{zhang2016, zhang2016b}.
Such data-driven models naturally allow for discontinuous operating modes by incorporating multiple
    feasible regions.
On the other hand, dynamic scale-bridging models can be trained from operating data, where
    transitions between feasible operating points are also modeled.
Recent works~\citep{pattison2016optimal, tsay2019optimal} accomplish this by performing system
    identification on (closed-loop) recorded process data, resulting in approximations of
    input–output relationships between process setpoints and outputs.
As the underlying variations in such data are few, data-mining techniques can further reduce the
    size of the dynamic scale-bridging models~\citep{tsay2020integrating}.
Overall, scale-bridging models derived using the above techniques can be used to compute optimal
    schedules of feasible operation.
\\

\textbf{Unsupervised learning.}
In terms of fault detection, process datasets are almost always highly unbalanced, i.e., nearly all
    recorded samples correspond to normal/routine operation.
Therefore, unsupervised learning is a well-established approach to use low variance, high volume data,
    e.g., anomaly detection, clustering, rather than supervised learning, e.g., explicit classification
    of faulty vs normal.
As in Section~\ref{sec:highvar}, dimensionality reduction, or manifold learning, seeks to represent
    high-dimensional data with a low-dimensional set of latent variables.
Note that supervised methods for dimensionality reduction also exist. 
Once the set of latent variables is learned, they can be used in fault detection~\citep{chiang2000fault},
    operational optimization~\citep{garcia2008optimization}, and process control~\citep{lauri2010data}.
\cite{macgregor2012monitoring} and \cite{qin2020bridging} discuss further applications and techniques.
Relatively simple fault detection rules can be derived by applying multivariate statistics, e.g.,
    Hotelling's $t$-squared statistic, to measured process data or latent variables~\citep{yin2014review}.
Clustering represents an alternative class of unsupervised learning, with the goal of partitioning data
    into a number of clusters based on some defined similarity metric.
By partitioning historical data into clusters, multiple operating modes of a process can be
    identified~\citep{quinones2019data} enabling, e.g., fault detection tailored to each operating mode.
Fault detection can also be constructed directly from a clustering scheme~\citep{detroja2006possibilistic},
    by observing how new data affect the learned clusters.
An advantage of using unsupervised techniques is the potential to detect new fault types, rather
    than only those used to train a supervised model.
\\

\textbf{Data-driven process control.} 
Machine learning techniques similar to the paradigm of ``reinforcement learning'' (RL) can use data
    to continually improve the performance of a process and its control system.
In contrast to the above methodologies, which primarily use available data to learn a model
    (from which control actions can be optimized), RL uses data to directly learn an optimal control
    policy~\citep{sutton2018reinforcement,hoskins1992process,shin2019reinforcement}.
However, the ``model-free'' RL approach is often more difficult, data-demanding, and therefore less
    effective for practical systems, in part due to the presence of noise~\citep{rawlings2019bringing}.
RL controllers may violate constraints while learning optimal policies; therefore, RL-based control
    is also popular in fields with fewer safety-critical constraints, e.g., building
    systems~\citep{wang2020reinforcement}.
\review{Even when using a dynamic process model is feasible and desirable, an approximate explicit
    MPC controller can still be trained based on closed-loop process
    data~\citep{aakesson2006neural,hussain1999review,lovelett2020some}.}
A major challenge here is incorporating system and controller constraints in the learned control
    policy~\citep{vaupel2020accelerating}.
Finally, a direct way to deal with low variations in process data is to enforce variations via
    future control actions.
Here, ``dual control,'' or simultaneous identification and control, can be achieved by heuristically
    incorporating system excitation in a process control problem~\citep{mesbah2018stochastic}.
\textit{Persistent excitation} strategies add constraints to the standard MPC problem to maintain a
    minimum level of excitation~\citep{genceli1996new}.
System excitation can also be enforced as a secondary objective, resulting in a multi-objective MPC
    problem~\citep{aggelogiannaki2006multiobjective, feng2018real, heirung2015mpc}

\section{Noisy/corrupt/missing data}
\label{sec:noisy}

\begin{figure}
    \centering
    \includegraphics[width=0.7\textwidth]{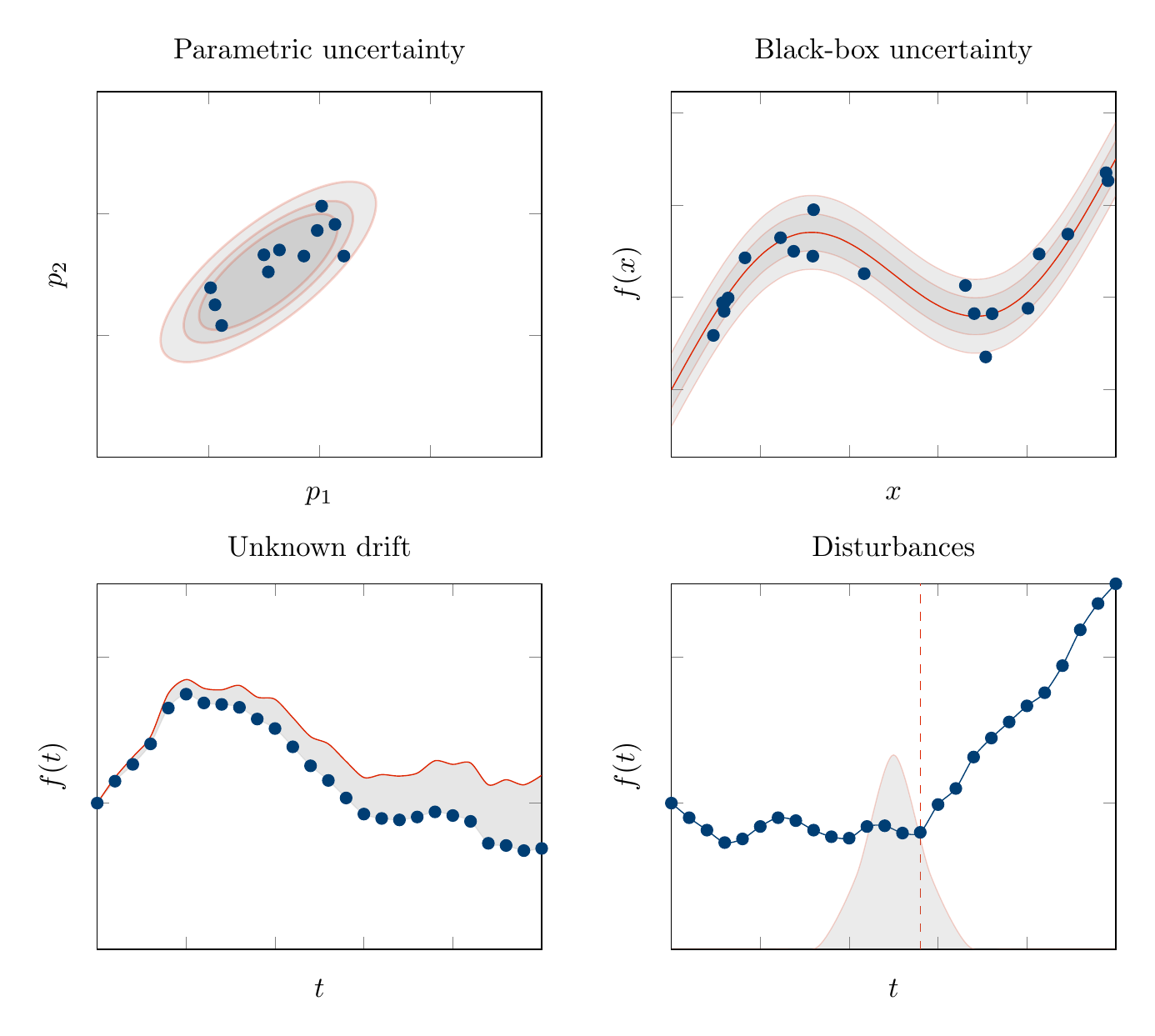}
    \caption{Uncertainty in chemical engineering data sets}
    \label{fig:uncertainty}
\end{figure}

Chemical engineering data sets are often noisy and contain corrupted or missing values, so
    applying machine learning methods requires considering these properties.
Figure~\ref{fig:uncertainty} illustrates the types of uncertainty which can arise from low quality data.
This section discusses how noisy, corrupted, and missing data arise in chemical engineering
    applications, where and why this type of data set is particularly relevant, and how machine
    learning has been applied to low quality data.
Next to noise and corruption in the data, we consider two different types of missing data.
First, data sets may contain missing values because of hardware failures, limited sensor
    ranges, or outlier removal.
Second, data which is relevant to the process may simply not be in the data set, because sensors
    could not be installed due to economic or physical constraints.
Different methods may be applicable to data sets with missing data depending on whether values are
    missing due to hardware issues or economic and physical constraints.

\subsection{Relevant applications}

The following discusses several applications where noisy, corrupted, and missing data are
    particularly relevant and have been addressed in detail in the literature. \\

\textbf{Fault detection.}
Section~\ref{sec:lowvar} discusses machine learning for fault detection, but process data used for
    detecting and pin-pointing equipment failures are also subject to noisy, corrupt, and missing data.
Fault detection has been studied in the presence of both missing data~\citep{He2009,Zhang2014,Askarian2016,Guo2020a}
    and noisy data~\citep{Venkatasubramanian1989,Hoang2019,Pham2020}.
Major challenges in applying fault detection to this type of data sets are:
    (1) developing methods which work in the presence of missing values, (2) ensuring noisy data do
    not lead to false positive faults, and (3) ensuring noisy or corrupted data do not lead to
    delays or failures to detect faults.
Addressing these challenges is important to assure the accuracy of fault detection and therefore
    the reliability of the plant. \\

\textbf{Degradation and fouling modeling.}
Equipment degradation and fouling is increasingly of concern not just in condition-monitoring,
    but also process design, planning, and scheduling~\citep{Yildirim2017,Basciftci2018,Wiebe2018,Wiebe2020}.
Degradation is usually modeled by stochastic models because of its inherent randomness and the
    limitations of vibrational and other data underlying these models~\citep{Jardine2006}.
Making use of such models in a process design, monitoring, or control context requires careful
    consideration of the data limitations and resulting uncertainties.
Challenges include: (1) developing models capturing the underlying equipment degradation in the
    presence of noisy/corrupted data and (2) making decisions based on uncertain equipment
    degradation states. \\

\textbf{State estimation and soft-sensing.}
Section~\ref{sec:lowvar} discusses state estimation as a method for dealing with missing data
    due to economical or physical restrictions, but these methods themselves can also be subject
    to errors in the input process data~\citep{Kadlec2011,Wang2018,Guo2020b}.
A main challenge arising from this is developing state estimation and soft-sensing methods
    which are robust to missing or noisy input data.
Other control-related areas relevant to low quality data include disturbance detection and
    rejection~\citep{Lawrynczuk2008}, sensor planning~\citep{Tewari2020}, trend estimation or slow
    feature analysis~\citep{Zhao2018,Si2019}, meta-learning and sparse optimization/compressed sensing.

\subsection{How this type of data is addressed in the literature}
Chemical engineering contributions have taken two different perspectives on
    low quality data, both of which routinely employ machine learning techniques.
The \textit{data perspective} considers noise, corruption,
    and missing values to be data properties which need to be addressed.
Such approaches often try to pre-process the data or develop methods which are robust, e.g., to missing values.
On the other hand, the \textit{uncertainty perspective} recognizes that noise, corruption,
    and missing values lead to uncertainty in process knowledge.
Instead of trying to ``fix'' the data, approaches that take the uncertainty view acknowledge the
    existence of uncertainty and try to make good decisions despite it.
Noisy measurements can lead to both parametric uncertainty, if a process parameter is measured,
    or to black-box uncertainty, if the data represent an unknown functional dependency.
Missing values can exacerbate this uncertainty.
Systematic corruption, e.g., by sensor drift or coking, can lead to measurement drifts while
    process conditions remain constant.
Section~\ref{sec:lowvar} discusses process drift in more detail.
The time at which disturbances or equipment failures occur may also be uncertain due to noisy
    measurements or missing values.
Below, we discuss the main approaches for using machine learning to deal with noisy, corrupted, or
    missing data from both perspectives.

While low quality data are an important source of uncertainty in chemical
    engineering processes, there are other sources of uncertainty as well.
For example the small size of available datasets can also be an important source of
    uncertainty, as discussed in Section~\ref{sec:highvar}.\\

\textbf{Data pre-processing.}
The most common approach for addressing low quality data from the data perspective is pre-processing~\citep{Xu2015}.
In the pre-processing paradigm, missing values are usually addressed by data imputation.
Missing data imputation for process engineering data sets has been previously
    reviewed~\citep{Severson2017,Imtiaz2008,Walczak2001}.
For a detailed review of machine learning techniques for data imputation see \cite{Lin2020}.
Supervised learning is often applied to predict missing values, e.g., with $k$-nearest-neighbours,
    decision trees, random forests, and ANNs~\citep{Lin2020}.
Recent contributions in machine learning have also applied deep generative approaches such as
    generative adversarial networks (GANs) or variational autoencoders (VAEs) for data
    imputation~\citep{Yoon2018,Camino2019,Nazabal2020}.
These approaches can achieve high accuracy but tend to have many parameters and therefore require
    large training sets.
They may therefore not be applicable to chemical engineering applications with low volume data regimes.
The choice of method most suitable for a given application depends on the amount of training data
    available, their quality, the percentage of missing values, and more.
Pre-processing for noisy and corrupted data includes outlier removal.
Both unsupervised machine learning approaches, e.g., $k$-means clustering~\citep{Pamula2011}, and
    supervised approaches, e.g., support vector machines, have been used to detect and remove outliers.
Unlike supervised approaches, clustering based approaches do not require a training set with labeled
    outliers.
\\

\textbf{Models which are robust to noise/missing values.}
An alternative approach to pre-processing is to use techniques which work in the presence of and
    are robust to noisy, corrupted, and missing data.
Many machine learning methods cannot be applied when the data contain missing values.
To alleviate this, \cite{Eirola2013,Eirola2014} propose distance estimation between vectors with
    missing values which allows distance-based machine learning algorithms like $k$-nearest-neighbours
    or support vector machines to be applied to data sets with missing values without pre-processing.
\cite{Mesquita2019} propose a similar approach for estimating the expected
    value of Gaussian kernels with incomplete data.
These techniques have the advantage that they can also be applied when a large percentage of
    values is missing.
Other approaches which have been successfully applied in the presence of missing and noisy data are
    Bayesian networks~\citep{Zhang2014,Askarian2016}.
While most machine learning techniques can be applied without modification in the presence of noise,
    noise may lead to inaccurate predictions or overfitting.
\\

\textbf{Stochastic models}
The uncertainty perspective to low quality data often starts by modeling the data using a stochastic model.
Stochastic models are often a better choice than deterministic models for noisy or corrupted data
    sets because they quantify the resulting uncertainty.
Stochastic processes can be interpreted as probability distributions over functions.
As such, they are particularly useful for modeling uncertainty in functional dependencies based on
    noisy or corrupted data, i.e., black-box function uncertainty.
Simple L\'evy type stochastic processes, like the Wiener or Gamma process, are commonly
    used in data-based condition monitoring/equipment degradation~\citep{Zhang2015,Nguyen2018}.
These models have also increasingly been incorporated into process scheduling and planning
    applications~\citep{Wiebe2018}.
Another commonly used class of stochastic processes, Gaussian processes (GPs), are extensively used
    as surrogate models in chemical engineering, but have also been used for applications with
    black-box uncertainty due to noisy measurements of functional dependencies~\citep{Wiebe2020,Liu2020,Bradford2020}.
Other stochastic models which have been applied to this type of data include Dirichlet process and
    Gaussian mixture models~\citep{Campbell2015,Chen2010,ning2018ddsro}, as well as Bayesian
    networks~\citep{Zhang2015,Jain2018}.
    \\

\textbf{Data-driven optimization under uncertainty}.
Several recent contributions in process systems engineering use data-driven optimization under
    uncertainty to combine machine learning with robust optimization to make optimal
    decisions based on noisy or corrupted data.
Data-driven robust optimization uses data to construct uncertainty sets.
Constraints where uncertain parameters occur are then required to hold for all values of the
    parameter within the uncertainty set.
Early approaches in data-driven robust optimization focused on constructing uncertainty sets from
    data using confidence regions and statistical hypothesis testing~\citep{Bertsimas2018}.
Other authors use Dirichlet process mixture models to construct unions of ellipsoidal uncertainty
    sets~\citep{Campbell2015} or polyhedral uncertainty sets~\citep{ning2018ddsro}.
The Dirichlet approach has the advantage that it can capture multimodal uncertainties.
It may therefore be particularly applicable to chemical engineering data sets where
    multimodal uncertainty is common, e.g., due to distinct operating modes.
Other contributions use unsupervised learning to construct data-driven
    uncertainty sets, e.g., \cite{shang2017svc} use kernel-based support vector clustering to
    derive data-driven uncertainty sets while \cite{Goerigk2020} construct sets from the output of
    an unsupervised deep neural networks.
ANNs have also been used in the context of distributionally robust and chance constraint optimization.
\cite{Zhao2020} use GANs to construct empirical distributions based on (noisy) data and create
    ambiguity sets for distributionally robust optimization based on these distribution.
While robust optimization is traditionally used for addressing parametric uncertainty, it can also
    be applied to black-box function uncertainty, e.g., \cite{Wiebe2020} use GPs to model black-box
    functions based on data and develop a method for robust reformulation of
    constraints depending on these black-box functions.\\

\section{Restricted data}
\label{sec:restricted}
Advances in sensor technologies and process monitoring/control systems have made large amounts of
    process data easily accessible for chemical processes.
Emerging technologies, such as Internet of Things~\citep{atzori2010internet} and
    Industry 4.0~\citep{lasi2014industry}, are also promising more efficient data collection
    and integration in the process industry~\citep{isaksson2018impact}.
However, there are fundamental challenges in data acquisition for chemical processes that can
    make it impossible to sample or measure properties during some operating conditions,
    resulting in restricted data sets.
Restricted data sets can cause difficulties in directly applying classical data-based technologies.

\subsection{Applications dealing with restricted data}
There are unique challenges in applying machine learning methodologies in chemical engineering
    that arise from restricted data sets.
This section briefly describes application areas and discusses challenges due to restricted data sets.\\

\textbf{Process control.} \review{Machine learning for process control has been an active research
    topic since the early 1990s~\citep{hoskins1992process,bhat1990use}, and this topic has gained
    more attention in the last years due to advances in machine
    learning~\citep{shin2019reinforcement,rawlings2019bringing}.}
\review{The restricted data setting creates several challenges for process control including:
    (1) learning accurate dynamic models from restricted data sets, (2) incorporating knowledge
    of hidden constraints, (3) reliably solving the large-scale constrained nonlinear optimization
    problems formed by MPC, (4) learning a control law within the process limitations,
    (5) ensuring a controller respects limitations and safety requirements at all time, and
    (6) identifying model mismatch.} \\

\textbf{Optimizing operation and production processes from data.}
Accurate mathematical models may not be known for each process unit, and some properties can
    change continuously or require mathematical relations too complicated to be directly integrated
    into an optimization problem.
By constructing data-driven surrogate models, it is possible to form a mathematical model of the
    optimization tasks containing some unknown relations.
Applications learning a model from data and using the model within an optimization framework for
    decision-making include: flowsheet optimization~\citep{caballero2008algorithm},
    superstructure optimization~\citep{henao2011surrogate}, supply chain
    management~\citep{wan2005simulation}, and process intensification~\citep{gutierrez2015multiobjective,quirante2015optimization}.
Some challenges within the restricted data setting are: (1) training accurate models on the
    restricted data sets, (2) learning constraints for the problem, (3) learning models of
    appropriate complexity and difficulty for the optimization problem, (4) incorporating model
    uncertainty, and (5) solving the resulting optimization problems.\\

\textbf{Inverse problems and product discovery.}
As previously mentioned, chemical engineering applications of inverse problems include product
    discovery and design of materials.
Restricted data sets also create challenges for inverse problems.
For example, data might only be available for certain regions of the input space, resulting in
    models with high uncertainty in large parts of the input space.
In molecular design, constraints arise from structural constraints, chemical feasibility, and
    required product properties~\citep{harper2000multi,austin2016computer,folic2008computer,gani2004computer}.
Unstable molecular designs may appear as hidden constraints in the inverse problem.
The challenges include: (1) learning accurate models from restricted data sets,
    (2) taking restrictions and hidden constraints into consideration, (3) incorporating model
    accuracy and uncertainty into  the inverse problem, and (4) efficiently solving the resulting
    optimization/inverse problem.

\subsection{How restricted data challenges are addressed in the literature}
This section reviews existing approaches for addressing some of the challenges of restricted data.
These are active research topics, and there might exist multiple solutions to these challenges. \\

\textbf{Learn constraints and limitations from data.}
Identifying feasible regions is a key component in flexibility
    analysis~\citep{swaney1985index,grossmann2014evolution}, but the classical approaches assume
    that algebraic constraints containing some uncertain parameters are
    known~\citep{halemane1983optimal,grossmann1987active}.
We refer the interested reader to \cite{banerjee2010computationally}, \cite{rogers2015feasibility},
    \cite{wang2017novel}, and \cite{metta2020novel} for reviews covering approaches to estimate
    feasible regions and constraints based on surrogate functions.
However, to efficiently model constraints with surrogate functions requires data samples in all
    regions of interests and a mixture of data points where the constraints are both
    satisfied and violated.
Especially in situations where data must be collected during  normal operation, there may be very
    few (if any) data points available where the constraints are violated.
In such circumstances the surrogate function approach may not be a viable strategy.
One-class classification~\citep{ruff2018deep,khan2009survey} has been proposed to address similar
    situations in machine learning: one-class classification could also be used in chemical
    engineering applications to represent feasible process configurations, but the authors are not
    yet aware of any such applications.\\

\textbf{Dealing with hidden constraints.}
The main difficulty with hidden constraints is the inability to directly measure or observe the
    state or property behind the hidden constraint, i.e., we cannot quantify by how much the
    constraint is satisfied or violated.
If a hidden constraint is encountered within an optimization framework, we need to somehow
    acknowledge that the point is infeasible and move away from the infeasible solution.
But, due to the hidden nature of the constraint, we might not obtain any other information than
    that the current point, e.g., process configuration, is infeasible.
With derivative free optimization algorithms, such as population-based search methods, hidden
    constraints can be directly dealt with by a simple penalty approach~\citep{martelli2014pgs},
    where a large penalty is imposed on the objective to force the search away from infeasible solutions.
Within an optimization framework, it could also be possible to use so-called, \textit{no-good}
    cuts~\citep{nannicini2012rounding,d2010interval} to exclude a small neighborhood around the
    infeasible point from the search space.
No-good cuts adds complexity to the optimization problem, and a large number of such cuts may
    result in a computationally intractable optimization problem.
Another approach for dealing with hidden constraints is to use a support vector machine to identify
    and remove infeasible solutions from the search space~\citep{ibrahim2018optimization}.\\

\textbf{Learning models from restricted data sets.}
The challenges in learning accurate models from restricted data sets can be quite similar to
    learning from small data sets.
Transfer learning~\citep{pan2009survey,taylor2007cross} is a machine learning concept that reuses
    knowledge learned from a similar task to improve performance and reduce the amounts of
    training data needed.
Transfer learning could also be useful for efficiently learning models from restricted data sets
    for chemical engineering applications.
For example, consider creating a model that predicts the product yield for a chemical process from a
    set of operating conditions.
Training an accurate model might require a large data set with high variance, but using knowledge
    and data from a similar process might greatly reduce the amount of data needed.
A simple approach to practically implement transfer learning for such applications is to use
    (part of) a model trained for a similar task as a starting point for the new model, e.g.,
    by reusing weights from some layers of an ANN.
Hybrid ANNs can be another approach for dealing with small and restricted data sets by incorporating
    known physical relations or first-principles equations, reducing the complexity of the model
    that is leaned from data~\citep{psichogios1992hybrid,medsker2012hybrid,bellos2005modelling}.
For example, physical relations can be incorporated by penalizing physical inconsistencies in the
    loss function while training ANNs~\citep{raissi2019physics,raissi2018deep,karpatne2017physics}.
There is also a risk of overfitting the models on restricted data sets, resulting in models with
    overall poor performance.
Several methods for training low complexity and sparse models have been
    presented~\citep{wilson2017alamo,bishop2006pattern, louizos2017learning,manngaard2018structural},
    which can improve the generalization ability of models trained on restricted data sets. \\

\textbf{Safety guarantees.}
The black-box nature of many machine learning models creates a set of challenges regarding safety
    guarantees, especially for automatic control applications.
For many control applications, it is crucial that the controller behaves as expected under all
    circumstance to avoid dangerous situations.
Therefore, successfull implementation of a data-driven controller in a safety-critical application
    might requires some safety guarantees.
For example, we could ideally ensure that the controller applies reasonable control actions under
    all circumstances and does not take \say{forbidden} control actions.
Adversarial examples for image classification~\citep{goodfellow2014explaining} have highlighted the
    sensitivity of ANNs by showing examples where image classifications can drastically change by
    practically invisible perturbations to the input images.
This has led to the development of robust verification
    techniques~\citep{cheng2017maximum,bunel2018unified,botoeva2020efficient, ehlers2017formal},
    that analyze the input-output behaviour of the ANNs by proving if certain outputs can occur
    while the inputs are restricted to specific domains~\citep{carlini2017towards}.
\review{Optimization and verification techniques might also be useful for obtaining safety guarantees
    for ANNs in process control applications.
For example, \citet{dai2021lyapunov} guarantee Lyapunov stability of ANN controllers during training
    by also learning a Lyapunov function as an ANN, while \citet{paulson2020approximate} propose a
    projection operator for guaranteeing feasibility and constraint satisfaction.} \\

\textbf{Data-driven process optimization.}
Here we consider situations where mathematical models are known for some processes and unit
    operations, but others are not fully known.
To solve the optimization task, we need to combine both known algebraic expressions with learned
    surrogate models into a tractable optimization problem.
Several types of surrogate models have been used for chemical engineering applications,
    including from ANNs~\citep{schweidtmann2019deterministic}, radial basis functions~\citep{wang2017novel},
    and Kriging/Gaussian processes~\citep{palmer2002metamodeling,caballero2008algorithm,jia2009predictive}.
Restrictions with known algebraic expressions can directly be incorporated into the optimization
    problem, whereas unknown restrictions might also need to be learned from data and represented
    using surrogate models.
\citet{eason2016trust} presented a trust-region filter method for this specific problem structure.
\citet{Bhosekar2018} provide an overview of surrogate-based optimization methods.
The optimization task here can also be viewed as a constrained black-box optimization problem,
    where the process units/operations with unknown mathematical models are considered as
    black-box functions.
A variety of methods hav ebeen proposed for constrained black-box optimization
    problems~\citep{audet2006mesh,hernandez2016general,conn2009introduction,banks2008review,boukouvala2016global}.
We refer the interested reader to \citet{conn2009introduction,boukouvala2016global} for more
    details on black-box optimization.\\

\textbf{Optimizing over ML models with constraints}.
Most of the applications mentioned in this section involve optimizing over ML models with constraints
    that represent various limitations.
Solving these optimization problems efficiently can be challenging, and there are different
    approaches available depending on the type of machine learning model.
Some machine learning models, such as ANNs with piecewise linear activation functions, can directly
    be represented as mixed integer linear optimization
    problems~\citep{anderson2020strong,fischetti2018deep, tsay2021partition}, and solved by
    established software.
Global optimization techniques have also been presented for
    gradient boosted trees~\citep{mistry2020mixed,thebelt2020entmoot} and neural networks with
    sigmodial activation functions~\citep{schweidtmann2019deterministic}.
If the entire machine learning model and optimization problem can be represented by algebraic
    equations and inequalities, the problem can be passed to a deterministic global optimization solver,
    such as ANTIGONE~\citep{misener2014antigone}, BARON~\citep {tawarmalani2005polyhedral},
    or SCIP~\citep{gamrath2020scip}.

\section{Conclusion}
The overall vision of this review paper is the observation that chemical engineers are using
    (1) traditional engineering approaches, (2) classical artificial intelligence,
    and (3) new research at the intersection of chemical engineering and artificial intelligence
    to derive significant value from ``information-poor'' data.
The commonality between each of the four data characteristics we identify is that each of these
    four types of data would be unsuited to classical machine learning approaches: the challenge
    for researchers at the interface between chemical engineering and computer science is to
    increase the information gained from the resulting available data.

\section*{Acknowledgements}

\noindent
During this project, A.T.\ was supported by BASF SE, Ludwigshafen am Rhein. J.W.\ was funded by
    the Engineering \& Physical Sciences Research Council (EPSRC) Centre for Doctoral Training in
    High Performance Embedded \& Distributed Systems (EP/L016796/1), and  an EPSRC/Schlumberger CASE
    studentship (EP/R511961/1, voucher 17000145). J.K.\ was funded by a Newton International
    Fellowship by the Royal Society (NIF\textbackslash R1\textbackslash 182194) and by the
    Swedish Cultural Foundation in Finland. R.M.\ and C.T.\ were funded by EPSRC Fellowships
    (grant numbers EP/P016871/1 and EP/T001577/1). C.T.\ also acknowledges support from an
    Imperial College Research Fellowship.

\bibliographystyle{elsarticle-harv}
\bibliography{lit.bib}

\begin{thebibliography}{208}
\expandafter\ifx\csname natexlab\endcsname\relax\def\natexlab#1{#1}\fi
\providecommand{\url}[1]{\texttt{#1}}
\providecommand{\href}[2]{#2}
\providecommand{\path}[1]{#1}
\providecommand{\DOIprefix}{doi:}
\providecommand{\ArXivprefix}{arXiv:}
\providecommand{\URLprefix}{URL: }
\providecommand{\Pubmedprefix}{pmid:}
\providecommand{\doi}[1]{\href{http://dx.doi.org/#1}{\path{#1}}}
\providecommand{\Pubmed}[1]{\href{pmid:#1}{\path{#1}}}
\providecommand{\bibinfo}[2]{#2}
\ifx\xfnm\relax \def\xfnm[#1]{\unskip,\space#1}\fi
\bibitem[{Aggelogiannaki and
  Sarimveis(2006)}]{aggelogiannaki2006multiobjective}
\bibinfo{author}{Aggelogiannaki, E.}, \bibinfo{author}{Sarimveis, H.},
  \bibinfo{year}{2006}.
\newblock \bibinfo{title}{Multiobjective constrained {MPC} with simultaneous
  closed-loop identification}.
\newblock \bibinfo{journal}{International Journal of Adaptive Control and
  Signal Processing} \bibinfo{volume}{20}, \bibinfo{pages}{145--173}.
\bibitem[{{\AA}kesson and Toivonen(2006)}]{aakesson2006neural}
\bibinfo{author}{{\AA}kesson, B.M.}, \bibinfo{author}{Toivonen, H.T.},
  \bibinfo{year}{2006}.
\newblock \bibinfo{title}{A neural network model predictive controller}.
\newblock \bibinfo{journal}{Journal of Process Control} \bibinfo{volume}{16},
  \bibinfo{pages}{937--946}.
\bibitem[{Alessandri et~al.(2011)Alessandri, Baglietto, Battistelli and
  Gaggero}]{alessandri2011moving}
\bibinfo{author}{Alessandri, A.}, \bibinfo{author}{Baglietto, M.},
  \bibinfo{author}{Battistelli, G.}, \bibinfo{author}{Gaggero, M.},
  \bibinfo{year}{2011}.
\newblock \bibinfo{title}{Moving-horizon state estimation for nonlinear systems
  using neural networks}.
\newblock \bibinfo{journal}{IEEE Transactions on Neural Networks}
  \bibinfo{volume}{22}, \bibinfo{pages}{768--780}.
\bibitem[{Anderson et~al.(2020)Anderson, Huchette, Ma, Tjandraatmadja and
  Vielma}]{anderson2020strong}
\bibinfo{author}{Anderson, R.}, \bibinfo{author}{Huchette, J.},
  \bibinfo{author}{Ma, W.}, \bibinfo{author}{Tjandraatmadja, C.},
  \bibinfo{author}{Vielma, J.P.}, \bibinfo{year}{2020}.
\newblock \bibinfo{title}{Strong mixed-integer programming formulations for
  trained neural networks}.
\newblock \bibinfo{journal}{Mathematical Programming} , \bibinfo{pages}{1--37}.
\bibitem[{Askarian et~al.(2016)Askarian, Escudero, Graells, Zarghami,
  Jalali-Farahani and Mostoufi}]{Askarian2016}
\bibinfo{author}{Askarian, M.}, \bibinfo{author}{Escudero, G.},
  \bibinfo{author}{Graells, M.}, \bibinfo{author}{Zarghami, R.},
  \bibinfo{author}{Jalali-Farahani, F.}, \bibinfo{author}{Mostoufi, N.},
  \bibinfo{year}{2016}.
\newblock \bibinfo{title}{Fault diagnosis of chemical processes with incomplete
  observations: A comparative study}.
\newblock \bibinfo{journal}{Computers \& Chemical Engineering}
  \bibinfo{volume}{84}, \bibinfo{pages}{104--116}.
\bibitem[{Atzori et~al.(2010)Atzori, Iera and Morabito}]{atzori2010internet}
\bibinfo{author}{Atzori, L.}, \bibinfo{author}{Iera, A.},
  \bibinfo{author}{Morabito, G.}, \bibinfo{year}{2010}.
\newblock \bibinfo{title}{The internet of things: A survey}.
\newblock \bibinfo{journal}{Computer networks} \bibinfo{volume}{54},
  \bibinfo{pages}{2787--2805}.
\bibitem[{Audet and Dennis~Jr(2006)}]{audet2006mesh}
\bibinfo{author}{Audet, C.}, \bibinfo{author}{Dennis~Jr, J.E.},
  \bibinfo{year}{2006}.
\newblock \bibinfo{title}{Mesh adaptive direct search algorithms for
  constrained optimization}.
\newblock \bibinfo{journal}{SIAM Journal on Optimization} \bibinfo{volume}{17},
  \bibinfo{pages}{188--217}.
\bibitem[{Austin et~al.(2016)Austin, Sahinidis and Trahan}]{austin2016computer}
\bibinfo{author}{Austin, N.D.}, \bibinfo{author}{Sahinidis, N.V.},
  \bibinfo{author}{Trahan, D.W.}, \bibinfo{year}{2016}.
\newblock \bibinfo{title}{Computer-aided molecular design: An introduction and
  review of tools, applications, and solution techniques}.
\newblock \bibinfo{journal}{Chemical Engineering Research \& Design}
  \bibinfo{volume}{116}, \bibinfo{pages}{2--26}.
\bibitem[{Banerjee et~al.(2010)Banerjee, Pal and
  Maiti}]{banerjee2010computationally}
\bibinfo{author}{Banerjee, I.}, \bibinfo{author}{Pal, S.},
  \bibinfo{author}{Maiti, S.}, \bibinfo{year}{2010}.
\newblock \bibinfo{title}{Computationally efficient black-box modeling for
  feasibility analysis}.
\newblock \bibinfo{journal}{Computers \& Chemical Engineering}
  \bibinfo{volume}{34}, \bibinfo{pages}{1515--1521}.
\bibitem[{Banks et~al.(2008)Banks, Vincent and Anyakoha}]{banks2008review}
\bibinfo{author}{Banks, A.}, \bibinfo{author}{Vincent, J.},
  \bibinfo{author}{Anyakoha, C.}, \bibinfo{year}{2008}.
\newblock \bibinfo{title}{A review of particle swarm optimization. part ii:
  hybridisation, combinatorial, multicriteria and constrained optimization, and
  indicative applications}.
\newblock \bibinfo{journal}{Natural Computing} \bibinfo{volume}{7},
  \bibinfo{pages}{109--124}.
\bibitem[{Bart{\'o}k et~al.(2013)Bart{\'o}k, Kondor and
  Cs{\'a}nyi}]{bartok2013representing}
\bibinfo{author}{Bart{\'o}k, A.P.}, \bibinfo{author}{Kondor, R.},
  \bibinfo{author}{Cs{\'a}nyi, G.}, \bibinfo{year}{2013}.
\newblock \bibinfo{title}{On representing chemical environments}.
\newblock \bibinfo{journal}{Physical Review B} \bibinfo{volume}{87},
  \bibinfo{pages}{184115}.
\bibitem[{Basciftci et~al.(2018)Basciftci, Ahmed, Gebraeel and
  Yildirim}]{Basciftci2018}
\bibinfo{author}{Basciftci, B.}, \bibinfo{author}{Ahmed, S.},
  \bibinfo{author}{Gebraeel, N.Z.}, \bibinfo{author}{Yildirim, M.},
  \bibinfo{year}{2018}.
\newblock \bibinfo{title}{Stochastic optimization of maintenance and operations
  schedules under unexpected failures}.
\newblock \bibinfo{journal}{IEEE Transactions on Power Systems}
  \bibinfo{volume}{8950}, \bibinfo{pages}{1--1}.
\bibitem[{Bellos et~al.(2005)Bellos, Kallinikos, Gounaris and
  Papayannakos}]{bellos2005modelling}
\bibinfo{author}{Bellos, G.}, \bibinfo{author}{Kallinikos, L.},
  \bibinfo{author}{Gounaris, C.}, \bibinfo{author}{Papayannakos, N.},
  \bibinfo{year}{2005}.
\newblock \bibinfo{title}{Modelling of the performance of industrial hds
  reactors using a hybrid neural network approach}.
\newblock \bibinfo{journal}{Chemical Engineering and Processing: Process
  Intensification} \bibinfo{volume}{44}, \bibinfo{pages}{505--515}.
\bibitem[{Bemporad et~al.(1999)Bemporad, Mignone and
  Morari}]{bemporad1999moving}
\bibinfo{author}{Bemporad, A.}, \bibinfo{author}{Mignone, D.},
  \bibinfo{author}{Morari, M.}, \bibinfo{year}{1999}.
\newblock \bibinfo{title}{Moving horizon estimation for hybrid systems and
  fault detection}, in: \bibinfo{booktitle}{Proceedings of the 1999 American
  Control Conference (Cat. No. 99CH36251)}, \bibinfo{organization}{IEEE}. pp.
  \bibinfo{pages}{2471--2475}.
\bibitem[{Bertsimas et~al.(2018)Bertsimas, Gupta and Kallus}]{Bertsimas2018}
\bibinfo{author}{Bertsimas, D.}, \bibinfo{author}{Gupta, V.},
  \bibinfo{author}{Kallus, N.}, \bibinfo{year}{2018}.
\newblock \bibinfo{title}{Data-driven robust optimization}.
\newblock \bibinfo{journal}{Mathematical Programming} \bibinfo{volume}{167},
  \bibinfo{pages}{235--292}.
\bibitem[{Bhagwat et~al.(2003)Bhagwat, Srinivasan and
  Krishnaswamy}]{bhagwat2003multi}
\bibinfo{author}{Bhagwat, A.}, \bibinfo{author}{Srinivasan, R.},
  \bibinfo{author}{Krishnaswamy, P.}, \bibinfo{year}{2003}.
\newblock \bibinfo{title}{Multi-linear model-based fault detection during
  process transitions}.
\newblock \bibinfo{journal}{Chemical Engineering Science} \bibinfo{volume}{58},
  \bibinfo{pages}{1649--1670}.
\bibitem[{Bhat and McAvoy(1990)}]{bhat1990use}
\bibinfo{author}{Bhat, N.}, \bibinfo{author}{McAvoy, T.J.},
  \bibinfo{year}{1990}.
\newblock \bibinfo{title}{Use of neural nets for dynamic modeling and control
  of chemical process systems}.
\newblock \bibinfo{journal}{Computers \& Chemical Engineering}
  \bibinfo{volume}{14}, \bibinfo{pages}{573--582}.
\bibitem[{Bhosekar and Ierapetritou(2018)}]{Bhosekar2018}
\bibinfo{author}{Bhosekar, A.}, \bibinfo{author}{Ierapetritou, M.},
  \bibinfo{year}{2018}.
\newblock \bibinfo{title}{Advances in surrogate based modeling, feasibility
  analysis, and optimization: A review}.
\newblock \bibinfo{journal}{Computers \& Chemical Engineering}
  \bibinfo{volume}{108}, \bibinfo{pages}{250--267}.
\bibitem[{Biegler et~al.(2014)Biegler, Lang and Lin}]{biegler2014multi}
\bibinfo{author}{Biegler, L.T.}, \bibinfo{author}{Lang, Y.d.},
  \bibinfo{author}{Lin, W.}, \bibinfo{year}{2014}.
\newblock \bibinfo{title}{Multi-scale optimization for process systems
  engineering}.
\newblock \bibinfo{journal}{Computers \& Chemical Engineering}
  \bibinfo{volume}{60}, \bibinfo{pages}{17--30}.
\bibitem[{Bishop(2006)}]{bishop2006pattern}
\bibinfo{author}{Bishop, C.M.}, \bibinfo{year}{2006}.
\newblock \bibinfo{title}{Pattern recognition and machine learning}.
\newblock \bibinfo{publisher}{Springer}.
\bibitem[{Botoeva et~al.(2020)Botoeva, Kouvaros, Kronqvist, Lomuscio and
  Misener}]{botoeva2020efficient}
\bibinfo{author}{Botoeva, E.}, \bibinfo{author}{Kouvaros, P.},
  \bibinfo{author}{Kronqvist, J.}, \bibinfo{author}{Lomuscio, A.},
  \bibinfo{author}{Misener, R.}, \bibinfo{year}{2020}.
\newblock \bibinfo{title}{Efficient verification of {ReLU}-based neural
  networks via dependency analysis.}, in: \bibinfo{booktitle}{AAAI}, pp.
  \bibinfo{pages}{3291--3299}.
\bibitem[{Boukouvala et~al.(2016)Boukouvala, Misener and
  Floudas}]{boukouvala2016global}
\bibinfo{author}{Boukouvala, F.}, \bibinfo{author}{Misener, R.},
  \bibinfo{author}{Floudas, C.A.}, \bibinfo{year}{2016}.
\newblock \bibinfo{title}{Global optimization advances in mixed-integer
  nonlinear programming, {MINLP}, and constrained derivative-free optimization,
  {CDFO}}.
\newblock \bibinfo{journal}{European Journal of Operational Research}
  \bibinfo{volume}{252}, \bibinfo{pages}{701--727}.
\bibitem[{Bradford et~al.(2020)Bradford, Imsland, Zhang and del
  Rio~Chanona}]{Bradford2020}
\bibinfo{author}{Bradford, E.}, \bibinfo{author}{Imsland, L.},
  \bibinfo{author}{Zhang, D.}, \bibinfo{author}{del Rio~Chanona, E.A.},
  \bibinfo{year}{2020}.
\newblock \bibinfo{title}{Stochastic data-driven model predictive control using
  {Gaussian} processes}.
\newblock \bibinfo{journal}{Computers \& Chemical Engineering}
  \bibinfo{volume}{139}.
\bibitem[{Bradford et~al.(2018a)Bradford, Schweidtmann and
  Lapkin}]{bradford2018efficient}
\bibinfo{author}{Bradford, E.}, \bibinfo{author}{Schweidtmann, A.M.},
  \bibinfo{author}{Lapkin, A.}, \bibinfo{year}{2018}a.
\newblock \bibinfo{title}{Efficient multiobjective optimization employing
  {Gaussian} processes, spectral sampling and a genetic algorithm}.
\newblock \bibinfo{journal}{Journal of Global Optimization}
  \bibinfo{volume}{71}, \bibinfo{pages}{407--438}.
\bibitem[{Bradford et~al.(2018b)Bradford, Schweidtmann, Zhang, Jing and del
  Rio-Chanona}]{bradford2018dynamic}
\bibinfo{author}{Bradford, E.}, \bibinfo{author}{Schweidtmann, A.M.},
  \bibinfo{author}{Zhang, D.}, \bibinfo{author}{Jing, K.}, \bibinfo{author}{del
  Rio-Chanona, E.A.}, \bibinfo{year}{2018}b.
\newblock \bibinfo{title}{Dynamic modeling and optimization of sustainable
  algal production with uncertainty using multivariate {Gaussian} processes}.
\newblock \bibinfo{journal}{Computers \& Chemical Engineering}
  \bibinfo{volume}{118}, \bibinfo{pages}{143--158}.
\bibitem[{Bunel et~al.(2018)Bunel, Turkaslan, Torr, Kohli and
  Mudigonda}]{bunel2018unified}
\bibinfo{author}{Bunel, R.R.}, \bibinfo{author}{Turkaslan, I.},
  \bibinfo{author}{Torr, P.}, \bibinfo{author}{Kohli, P.},
  \bibinfo{author}{Mudigonda, P.K.}, \bibinfo{year}{2018}.
\newblock \bibinfo{title}{A unified view of piecewise linear neural network
  verification}, in: \bibinfo{booktitle}{Advances in Neural Information
  Processing Systems}, pp. \bibinfo{pages}{4790--4799}.
\bibitem[{Caballero and Grossmann(2008)}]{caballero2008algorithm}
\bibinfo{author}{Caballero, J.A.}, \bibinfo{author}{Grossmann, I.E.},
  \bibinfo{year}{2008}.
\newblock \bibinfo{title}{An algorithm for the use of surrogate models in
  modular flowsheet optimization}.
\newblock \bibinfo{journal}{AIChE journal} \bibinfo{volume}{54},
  \bibinfo{pages}{2633--2650}.
\bibitem[{Camino et~al.(2019)Camino, Hammerschmidt and State}]{Camino2019}
\bibinfo{author}{Camino, R.D.}, \bibinfo{author}{Hammerschmidt, C.A.},
  \bibinfo{author}{State, R.}, \bibinfo{year}{2019}.
\newblock \bibinfo{title}{Improving missing data imputation with deep
  generative models}.
\newblock \bibinfo{journal}{arXiv preprint arXiv:1902.10666} .
\bibitem[{Campbell and How(2015)}]{Campbell2015}
\bibinfo{author}{Campbell, T.}, \bibinfo{author}{How, J.P.},
  \bibinfo{year}{2015}.
\newblock \bibinfo{title}{Bayesian nonparametric set construction for robust
  optimization}.
\newblock \bibinfo{journal}{Proceedings of the American Control Conference}
  \bibinfo{volume}{2015-July}, \bibinfo{pages}{4216--4221}.
\bibitem[{Carlini and Wagner(2017)}]{carlini2017towards}
\bibinfo{author}{Carlini, N.}, \bibinfo{author}{Wagner, D.},
  \bibinfo{year}{2017}.
\newblock \bibinfo{title}{Towards evaluating the robustness of neural
  networks}, in: \bibinfo{booktitle}{2017 ieee symposium on security and
  privacy (sp)}, \bibinfo{organization}{IEEE}. pp. \bibinfo{pages}{39--57}.
\bibitem[{Chen et~al.(2014)Chen, Fox and Guestrin}]{chen2014stochastic}
\bibinfo{author}{Chen, T.}, \bibinfo{author}{Fox, E.},
  \bibinfo{author}{Guestrin, C.}, \bibinfo{year}{2014}.
\newblock \bibinfo{title}{Stochastic gradient {Hamiltonian Monte Carlo}}, in:
  \bibinfo{booktitle}{International Conference on Machine Learning},
  \bibinfo{organization}{PMLR}. pp. \bibinfo{pages}{1683--1691}.
\bibitem[{Chen and Zhang(2010)}]{Chen2010}
\bibinfo{author}{Chen, T.}, \bibinfo{author}{Zhang, J.}, \bibinfo{year}{2010}.
\newblock \bibinfo{title}{On-line multivariate statistical monitoring of batch
  processes using gaussian mixture model}.
\newblock \bibinfo{journal}{Computers \& Chemical Engineering}
  \bibinfo{volume}{34}, \bibinfo{pages}{500--507}.
\bibitem[{Cheng et~al.(2017)Cheng, N{\"u}hrenberg and Ruess}]{cheng2017maximum}
\bibinfo{author}{Cheng, C.H.}, \bibinfo{author}{N{\"u}hrenberg, G.},
  \bibinfo{author}{Ruess, H.}, \bibinfo{year}{2017}.
\newblock \bibinfo{title}{Maximum resilience of artificial neural networks},
  in: \bibinfo{booktitle}{International Symposium on Automated Technology for
  Verification and Analysis}, \bibinfo{organization}{Springer}. pp.
  \bibinfo{pages}{251--268}.
\bibitem[{Chiang et~al.(2000)Chiang, Russell and Braatz}]{chiang2000fault}
\bibinfo{author}{Chiang, L.H.}, \bibinfo{author}{Russell, E.L.},
  \bibinfo{author}{Braatz, R.D.}, \bibinfo{year}{2000}.
\newblock \bibinfo{title}{Fault diagnosis in chemical processes using fisher
  discriminant analysis, discriminant partial least squares, and principal
  component analysis}.
\newblock \bibinfo{journal}{Chemometrics and intelligent laboratory systems}
  \bibinfo{volume}{50}, \bibinfo{pages}{243--252}.
\bibitem[{Clayton et~al.(2020)Clayton, Schweidtmann, Clemens, Manson, Taylor,
  Ni{\~n}o, Chamberlain, Kapur, Blacker, Lapkin et~al.}]{clayton2020automated}
\bibinfo{author}{Clayton, A.D.}, \bibinfo{author}{Schweidtmann, A.M.},
  \bibinfo{author}{Clemens, G.}, \bibinfo{author}{Manson, J.A.},
  \bibinfo{author}{Taylor, C.J.}, \bibinfo{author}{Ni{\~n}o, C.G.},
  \bibinfo{author}{Chamberlain, T.W.}, \bibinfo{author}{Kapur, N.},
  \bibinfo{author}{Blacker, A.J.}, \bibinfo{author}{Lapkin, A.A.}, et~al.,
  \bibinfo{year}{2020}.
\newblock \bibinfo{title}{Automated self-optimisation of multi-step reaction
  and separation processes using machine learning}.
\newblock \bibinfo{journal}{Chemical Engineering Journal}
  \bibinfo{volume}{384}, \bibinfo{pages}{123340}.
\bibitem[{Conn et~al.(2009)Conn, Scheinberg and Vicente}]{conn2009introduction}
\bibinfo{author}{Conn, A.R.}, \bibinfo{author}{Scheinberg, K.},
  \bibinfo{author}{Vicente, L.N.}, \bibinfo{year}{2009}.
\newblock \bibinfo{title}{Introduction to derivative-free optimization}.
\newblock \bibinfo{publisher}{SIAM}.
\bibitem[{Dai et~al.(2021)Dai, Landry, Yang, Pavone and
  Tedrake}]{dai2021lyapunov}
\bibinfo{author}{Dai, H.}, \bibinfo{author}{Landry, B.}, \bibinfo{author}{Yang,
  L.}, \bibinfo{author}{Pavone, M.}, \bibinfo{author}{Tedrake, R.},
  \bibinfo{year}{2021}.
\newblock \bibinfo{title}{Lyapunov-stable neural-network control}.
\newblock \bibinfo{journal}{arXiv preprint arXiv:2109.14152} .
\bibitem[{Daum(2005)}]{daum2005nonlinear}
\bibinfo{author}{Daum, F.}, \bibinfo{year}{2005}.
\newblock \bibinfo{title}{Nonlinear filters: beyond the {Kalman} filter}.
\newblock \bibinfo{journal}{IEEE Aerospace and Electronic Systems Magazine}
  \bibinfo{volume}{20}, \bibinfo{pages}{57--69}.
\bibitem[{Detroja et~al.(2006)Detroja, Gudi and
  Patwardhan}]{detroja2006possibilistic}
\bibinfo{author}{Detroja, K.}, \bibinfo{author}{Gudi, R.},
  \bibinfo{author}{Patwardhan, S.}, \bibinfo{year}{2006}.
\newblock \bibinfo{title}{A possibilistic clustering approach to novel fault
  detection and isolation}.
\newblock \bibinfo{journal}{Journal of Process Control} \bibinfo{volume}{16},
  \bibinfo{pages}{1055--1073}.
\bibitem[{Dong and Qin(2018)}]{dong2018dynamic}
\bibinfo{author}{Dong, Y.}, \bibinfo{author}{Qin, S.J.}, \bibinfo{year}{2018}.
\newblock \bibinfo{title}{Dynamic latent variable analytics for process
  operations and control}.
\newblock \bibinfo{journal}{Computers \& Chemical Engineering}
  \bibinfo{volume}{114}, \bibinfo{pages}{69--80}.
\bibitem[{D’Ambrosio et~al.(2010)D’Ambrosio, Frangioni, Liberti and
  Lodi}]{d2010interval}
\bibinfo{author}{D’Ambrosio, C.}, \bibinfo{author}{Frangioni, A.},
  \bibinfo{author}{Liberti, L.}, \bibinfo{author}{Lodi, A.},
  \bibinfo{year}{2010}.
\newblock \bibinfo{title}{On interval-subgradient and no-good cuts}.
\newblock \bibinfo{journal}{Operations Research Letters} \bibinfo{volume}{38},
  \bibinfo{pages}{341--345}.
\bibitem[{Eason and Biegler(2016)}]{eason2016trust}
\bibinfo{author}{Eason, J.P.}, \bibinfo{author}{Biegler, L.T.},
  \bibinfo{year}{2016}.
\newblock \bibinfo{title}{A trust region filter method for glass box/black box
  optimization}.
\newblock \bibinfo{journal}{AIChE Journal} \bibinfo{volume}{62},
  \bibinfo{pages}{3124--3136}.
\bibitem[{Ehlers(2017)}]{ehlers2017formal}
\bibinfo{author}{Ehlers, R.}, \bibinfo{year}{2017}.
\newblock \bibinfo{title}{Formal verification of piece-wise linear feed-forward
  neural networks}, in: \bibinfo{booktitle}{International Symposium on
  Automated Technology for Verification and Analysis},
  \bibinfo{organization}{Springer}. pp. \bibinfo{pages}{269--286}.
\bibitem[{Eirola et~al.(2013)Eirola, Doquire, Verleysen and
  Lendasse}]{Eirola2013}
\bibinfo{author}{Eirola, E.}, \bibinfo{author}{Doquire, G.},
  \bibinfo{author}{Verleysen, M.}, \bibinfo{author}{Lendasse, A.},
  \bibinfo{year}{2013}.
\newblock \bibinfo{title}{Distance estimation in numerical data sets with
  missing values}.
\newblock \bibinfo{journal}{Information Sciences} \bibinfo{volume}{240},
  \bibinfo{pages}{115--128}.
\bibitem[{Eirola et~al.(2014)Eirola, Lendasse, Vandewalle and
  Biernacki}]{Eirola2014}
\bibinfo{author}{Eirola, E.}, \bibinfo{author}{Lendasse, A.},
  \bibinfo{author}{Vandewalle, V.}, \bibinfo{author}{Biernacki, C.},
  \bibinfo{year}{2014}.
\newblock \bibinfo{title}{Mixture of {Gaussians} for distance estimation with
  missing data}.
\newblock \bibinfo{journal}{Neurocomputing} \bibinfo{volume}{131},
  \bibinfo{pages}{32--42}.
\bibitem[{Eklund et~al.(2014)Eklund, Norinder, Boyer and
  Carlsson}]{eklund2014choosing}
\bibinfo{author}{Eklund, M.}, \bibinfo{author}{Norinder, U.},
  \bibinfo{author}{Boyer, S.}, \bibinfo{author}{Carlsson, L.},
  \bibinfo{year}{2014}.
\newblock \bibinfo{title}{Choosing feature selection and learning algorithms in
  {QSAR}}.
\newblock \bibinfo{journal}{Journal of Chemical Information and Modeling}
  \bibinfo{volume}{54}, \bibinfo{pages}{837--843}.
\bibitem[{Esposito and Floudas(1998)}]{esposito1998global}
\bibinfo{author}{Esposito, W.R.}, \bibinfo{author}{Floudas, C.A.},
  \bibinfo{year}{1998}.
\newblock \bibinfo{title}{Global optimization in parameter estimation of
  nonlinear algebraic models via the error-in-variables approach}.
\newblock \bibinfo{journal}{Industrial \& Engineering Chemistry Research}
  \bibinfo{volume}{37}, \bibinfo{pages}{1841--1858}.
\bibitem[{Feng and Houska(2018)}]{feng2018real}
\bibinfo{author}{Feng, X.}, \bibinfo{author}{Houska, B.}, \bibinfo{year}{2018}.
\newblock \bibinfo{title}{Real-time algorithm for self-reflective model
  predictive control}.
\newblock \bibinfo{journal}{Journal of Process Control} \bibinfo{volume}{65},
  \bibinfo{pages}{68--77}.
\bibitem[{Fischetti and Jo(2018)}]{fischetti2018deep}
\bibinfo{author}{Fischetti, M.}, \bibinfo{author}{Jo, J.},
  \bibinfo{year}{2018}.
\newblock \bibinfo{title}{Deep neural networks and mixed integer linear
  optimization}.
\newblock \bibinfo{journal}{Constraints} \bibinfo{volume}{23},
  \bibinfo{pages}{296--309}.
\bibitem[{Folic et~al.(2008)Folic, Adjiman and
  Pistikopoulos}]{folic2008computer}
\bibinfo{author}{Folic, M.}, \bibinfo{author}{Adjiman, C.S.},
  \bibinfo{author}{Pistikopoulos, E.N.}, \bibinfo{year}{2008}.
\newblock \bibinfo{title}{Computer-aided solvent design for reactions:
  maximizing product formation}.
\newblock \bibinfo{journal}{Industrial \& Engineering Chemistry Research}
  \bibinfo{volume}{47}, \bibinfo{pages}{5190--5202}.
\bibitem[{Frazier(2018)}]{frazier2018tutorial}
\bibinfo{author}{Frazier, P.I.}, \bibinfo{year}{2018}.
\newblock \bibinfo{title}{A tutorial on {Bayesian} optimization}.
\newblock \bibinfo{journal}{arXiv preprint arXiv:1807.02811} .
\bibitem[{Gamrath et~al.(2020)Gamrath, Anderson, Bestuzheva, Chen, Eifler,
  Gasse, Gemander, Gleixner, Gottwald, Halbig et~al.}]{gamrath2020scip}
\bibinfo{author}{Gamrath, G.}, \bibinfo{author}{Anderson, D.},
  \bibinfo{author}{Bestuzheva, K.}, \bibinfo{author}{Chen, W.K.},
  \bibinfo{author}{Eifler, L.}, \bibinfo{author}{Gasse, M.},
  \bibinfo{author}{Gemander, P.}, \bibinfo{author}{Gleixner, A.},
  \bibinfo{author}{Gottwald, L.}, \bibinfo{author}{Halbig, K.}, et~al.,
  \bibinfo{year}{2020}.
\newblock \bibinfo{title}{The {SCIP} optimization suite 7.0} .
\bibitem[{Gani(2004)}]{gani2004computer}
\bibinfo{author}{Gani, R.}, \bibinfo{year}{2004}.
\newblock \bibinfo{title}{Computer-aided methods and tools for chemical product
  design}.
\newblock \bibinfo{journal}{Chemical Engineering Research \& Design}
  \bibinfo{volume}{82}, \bibinfo{pages}{1494--1504}.
\bibitem[{García-Muñoz et~al.(2008)García-Muñoz, MacGregor, Neogi, Latshaw
  and Mehta}]{garcia2008optimization}
\bibinfo{author}{García-Muñoz, S.}, \bibinfo{author}{MacGregor, J.F.},
  \bibinfo{author}{Neogi, D.}, \bibinfo{author}{Latshaw, B.E.},
  \bibinfo{author}{Mehta, S.}, \bibinfo{year}{2008}.
\newblock \bibinfo{title}{Optimization of batch operating policies. part ii.
  incorporating process constraints and industrial applications}.
\newblock \bibinfo{journal}{Industrial \& Engineering Chemistry Research}
  \bibinfo{volume}{47}, \bibinfo{pages}{4202--4208}.
\bibitem[{{García Muñoz} and Torres(2020)}]{doi:10.1021/acs.iecr.0c01385}
\bibinfo{author}{{García Muñoz}, S.}, \bibinfo{author}{Torres, E.H.},
  \bibinfo{year}{2020}.
\newblock \bibinfo{title}{Supervised extended iterative optimization technology
  for estimation of powder compositions in pharmaceutical applications: Method
  and lifecycle management}.
\newblock \bibinfo{journal}{Industrial \& Engineering Chemistry Research}
  \bibinfo{volume}{59}, \bibinfo{pages}{10072--10081}.
\newblock \URLprefix \url{https://doi.org/10.1021/acs.iecr.0c01385},
  \DOIprefix\doi{10.1021/acs.iecr.0c01385},
  \href{http://arxiv.org/abs/https://doi.org/10.1021/acs.iecr.0c01385}{{\tt
  arXiv:https://doi.org/10.1021/acs.iecr.0c01385}}.
\bibitem[{Garnelo et~al.(2018)Garnelo, Rosenbaum, Maddison, Ramalho, Saxton,
  Shanahan, Teh, Rezende and Eslami}]{garnelo2018conditional}
\bibinfo{author}{Garnelo, M.}, \bibinfo{author}{Rosenbaum, D.},
  \bibinfo{author}{Maddison, C.}, \bibinfo{author}{Ramalho, T.},
  \bibinfo{author}{Saxton, D.}, \bibinfo{author}{Shanahan, M.},
  \bibinfo{author}{Teh, Y.W.}, \bibinfo{author}{Rezende, D.},
  \bibinfo{author}{Eslami, S.A.}, \bibinfo{year}{2018}.
\newblock \bibinfo{title}{Conditional neural processes}, in:
  \bibinfo{booktitle}{International Conference on Machine Learning},
  \bibinfo{organization}{PMLR}. pp. \bibinfo{pages}{1704--1713}.
\bibitem[{Gau and Stadtherr(2002)}]{gau2002deterministic}
\bibinfo{author}{Gau, C.Y.}, \bibinfo{author}{Stadtherr, M.A.},
  \bibinfo{year}{2002}.
\newblock \bibinfo{title}{Deterministic global optimization for
  error-in-variables parameter estimation}.
\newblock \bibinfo{journal}{AIChE Journal} \bibinfo{volume}{48},
  \bibinfo{pages}{1192--1197}.
\bibitem[{Ge et~al.(2013)Ge, Song and Gao}]{ge2013review}
\bibinfo{author}{Ge, Z.}, \bibinfo{author}{Song, Z.}, \bibinfo{author}{Gao,
  F.}, \bibinfo{year}{2013}.
\newblock \bibinfo{title}{Review of recent research on data-based process
  monitoring}.
\newblock \bibinfo{journal}{Industrial \& Engineering Chemistry Research}
  \bibinfo{volume}{52}, \bibinfo{pages}{3543--3562}.
\bibitem[{Genceli and Nikolaou(1996)}]{genceli1996new}
\bibinfo{author}{Genceli, H.}, \bibinfo{author}{Nikolaou, M.},
  \bibinfo{year}{1996}.
\newblock \bibinfo{title}{New approach to constrained predictive control with
  simultaneous model identification}.
\newblock \bibinfo{journal}{AIChE journal} \bibinfo{volume}{42},
  \bibinfo{pages}{2857--2868}.
\bibitem[{Ghiringhelli et~al.(2015)Ghiringhelli, Vybiral, Levchenko, Draxl and
  Scheffler}]{ghiringhelli2015big}
\bibinfo{author}{Ghiringhelli, L.M.}, \bibinfo{author}{Vybiral, J.},
  \bibinfo{author}{Levchenko, S.V.}, \bibinfo{author}{Draxl, C.},
  \bibinfo{author}{Scheffler, M.}, \bibinfo{year}{2015}.
\newblock \bibinfo{title}{Big data of materials science: critical role of the
  descriptor}.
\newblock \bibinfo{journal}{Physical review letters} \bibinfo{volume}{114},
  \bibinfo{pages}{105503}.
\bibitem[{Goerigk and Kurtz(2020)}]{Goerigk2020}
\bibinfo{author}{Goerigk, M.}, \bibinfo{author}{Kurtz, J.},
  \bibinfo{year}{2020}.
\newblock \bibinfo{title}{Data-driven robust optimization using unsupervised
  deep learning}.
\newblock \href{http://arxiv.org/abs/2011.09769}{{\tt arXiv:2011.09769}}.
\bibitem[{Goodfellow et~al.(2014)Goodfellow, Shlens and
  Szegedy}]{goodfellow2014explaining}
\bibinfo{author}{Goodfellow, I.J.}, \bibinfo{author}{Shlens, J.},
  \bibinfo{author}{Szegedy, C.}, \bibinfo{year}{2014}.
\newblock \bibinfo{title}{Explaining and harnessing adversarial examples}.
\newblock \bibinfo{journal}{arXiv preprint arXiv:1412.6572} .
\bibitem[{G{\'o}rak and Sorensen(2014)}]{gorak2014distillation}
\bibinfo{editor}{G{\'o}rak, A.}, \bibinfo{editor}{Sorensen, E.} (Eds.),
  \bibinfo{year}{2014}.
\newblock \bibinfo{title}{Distillation: fundamentals and principles}.
\newblock \bibinfo{publisher}{Academic Press}.
\bibitem[{Green and Southard(2019)}]{green2019perry}
\bibinfo{author}{Green, D.W.}, \bibinfo{author}{Southard, M.Z.},
  \bibinfo{year}{2019}.
\newblock \bibinfo{title}{Perry's chemical engineers' handbook}.
\newblock \bibinfo{publisher}{McGraw-Hill Education}.
\bibitem[{Grossmann et~al.(2014)Grossmann, Calfa and
  Garcia-Herreros}]{grossmann2014evolution}
\bibinfo{author}{Grossmann, I.E.}, \bibinfo{author}{Calfa, B.A.},
  \bibinfo{author}{Garcia-Herreros, P.}, \bibinfo{year}{2014}.
\newblock \bibinfo{title}{Evolution of concepts and models for quantifying
  resiliency and flexibility of chemical processes}.
\newblock \bibinfo{journal}{Computers \& Chemical Engineering}
  \bibinfo{volume}{70}, \bibinfo{pages}{22--34}.
\bibitem[{Grossmann and Floudas(1987)}]{grossmann1987active}
\bibinfo{author}{Grossmann, I.E.}, \bibinfo{author}{Floudas, C.A.},
  \bibinfo{year}{1987}.
\newblock \bibinfo{title}{Active constraint strategy for flexibility analysis
  in chemical processes}.
\newblock \bibinfo{journal}{Computers \& Chemical Engineering}
  \bibinfo{volume}{11}, \bibinfo{pages}{675--693}.
\bibitem[{Guo et~al.(2020)Guo, Hu, Yang and Huang}]{Guo2020a}
\bibinfo{author}{Guo, C.}, \bibinfo{author}{Hu, W.}, \bibinfo{author}{Yang,
  F.}, \bibinfo{author}{Huang, D.}, \bibinfo{year}{2020}.
\newblock \bibinfo{title}{Deep learning technique for process fault detection
  and diagnosis in the presence of incomplete data}.
\newblock \bibinfo{journal}{Chinese Journal of Chemical Engineering} .
\bibitem[{Guo and Huang(2020)}]{Guo2020b}
\bibinfo{author}{Guo, F.}, \bibinfo{author}{Huang, B.}, \bibinfo{year}{2020}.
\newblock \bibinfo{title}{A mutual information-based variational autoencoder
  for robust jit soft sensing with abnormal observations}.
\newblock \bibinfo{journal}{Chemometrics and Intelligent Laboratory Systems}
  \bibinfo{volume}{204}.
\bibitem[{Guo(1990)}]{guo1990estimating}
\bibinfo{author}{Guo, L.}, \bibinfo{year}{1990}.
\newblock \bibinfo{title}{Estimating time-varying parameters by the kalman
  filter based algorithm: stability and convergence}.
\newblock \bibinfo{journal}{IEEE Transactions on Automatic Control}
  \bibinfo{volume}{35}, \bibinfo{pages}{141--147}.
\bibitem[{Guti{\'e}rrez-Antonio and
  Briones-Ram{\'\i}rez(2015)}]{gutierrez2015multiobjective}
\bibinfo{author}{Guti{\'e}rrez-Antonio, C.},
  \bibinfo{author}{Briones-Ram{\'\i}rez, A.}, \bibinfo{year}{2015}.
\newblock \bibinfo{title}{Multiobjective stochastic optimization of
  dividing-wall distillation columns using a surrogate model based on neural
  networks}.
\newblock \bibinfo{journal}{Chemical and biochemical engineering quarterly}
  \bibinfo{volume}{29}, \bibinfo{pages}{491--504}.
\bibitem[{Halemane and Grossmann(1983)}]{halemane1983optimal}
\bibinfo{author}{Halemane, K.P.}, \bibinfo{author}{Grossmann, I.E.},
  \bibinfo{year}{1983}.
\newblock \bibinfo{title}{Optimal process design under uncertainty}.
\newblock \bibinfo{journal}{AIChE Journal} \bibinfo{volume}{29},
  \bibinfo{pages}{425--433}.
\bibitem[{Harper and Gani(2000)}]{harper2000multi}
\bibinfo{author}{Harper, P.M.}, \bibinfo{author}{Gani, R.},
  \bibinfo{year}{2000}.
\newblock \bibinfo{title}{A multi-step and multi-level approach for computer
  aided molecular design}.
\newblock \bibinfo{journal}{Computers \& Chemical Engineering}
  \bibinfo{volume}{24}, \bibinfo{pages}{677--683}.
\bibitem[{Hashemian and Armaou(2015)}]{hashemian2015fast}
\bibinfo{author}{Hashemian, N.}, \bibinfo{author}{Armaou, A.},
  \bibinfo{year}{2015}.
\newblock \bibinfo{title}{Fast moving horizon estimation of nonlinear processes
  via carleman linearization}, in: \bibinfo{booktitle}{2015 American Control
  Conference (ACC)}, \bibinfo{organization}{IEEE}. pp.
  \bibinfo{pages}{3379--3385}.
\bibitem[{He et~al.(2009)He, Wang and Zhou}]{He2009}
\bibinfo{author}{He, X.}, \bibinfo{author}{Wang, Z.}, \bibinfo{author}{Zhou,
  D.H.}, \bibinfo{year}{2009}.
\newblock \bibinfo{title}{Robust fault detection for networked systems with
  communication delay and data missing}.
\newblock \bibinfo{journal}{Automatica} \bibinfo{volume}{45},
  \bibinfo{pages}{2634--2639}.
\bibitem[{Heirung et~al.(2015)Heirung, Foss and Ydstie}]{heirung2015mpc}
\bibinfo{author}{Heirung, T.A.N.}, \bibinfo{author}{Foss, B.},
  \bibinfo{author}{Ydstie, B.E.}, \bibinfo{year}{2015}.
\newblock \bibinfo{title}{{MPC}-based dual control with online experiment
  design}.
\newblock \bibinfo{journal}{Journal of Process Control} \bibinfo{volume}{32},
  \bibinfo{pages}{64--76}.
\bibitem[{Henao and Maravelias(2011)}]{henao2011surrogate}
\bibinfo{author}{Henao, C.A.}, \bibinfo{author}{Maravelias, C.T.},
  \bibinfo{year}{2011}.
\newblock \bibinfo{title}{Surrogate-based superstructure optimization
  framework}.
\newblock \bibinfo{journal}{AIChE Journal} \bibinfo{volume}{57},
  \bibinfo{pages}{1216--1232}.
\bibitem[{Herbol et~al.(2018)Herbol, Hu, Frazier, Clancy and
  Poloczek}]{herbol2018efficient}
\bibinfo{author}{Herbol, H.C.}, \bibinfo{author}{Hu, W.},
  \bibinfo{author}{Frazier, P.}, \bibinfo{author}{Clancy, P.},
  \bibinfo{author}{Poloczek, M.}, \bibinfo{year}{2018}.
\newblock \bibinfo{title}{Efficient search of compositional space for hybrid
  organic--inorganic perovskites via {Bayesian} optimization}.
\newblock \bibinfo{journal}{npj Computational Materials} \bibinfo{volume}{4},
  \bibinfo{pages}{1--7}.
\bibitem[{Hern{\'a}ndez-Lobato et~al.(2016)Hern{\'a}ndez-Lobato, Gelbart,
  Adams, Hoffman and Ghahramani}]{hernandez2016general}
\bibinfo{author}{Hern{\'a}ndez-Lobato, J.M.}, \bibinfo{author}{Gelbart, M.A.},
  \bibinfo{author}{Adams, R.P.}, \bibinfo{author}{Hoffman, M.W.},
  \bibinfo{author}{Ghahramani, Z.}, \bibinfo{year}{2016}.
\newblock \bibinfo{title}{A general framework for constrained {Bayesian}
  optimization using information-based search}.
\newblock \bibinfo{journal}{The Journal of Machine Learning Research}
  \bibinfo{volume}{17}, \bibinfo{pages}{5549--5601}.
\bibitem[{Hewing et~al.(2020)Hewing, Wabersich, Menner and
  Zeilinger}]{hewing2020learning}
\bibinfo{author}{Hewing, L.}, \bibinfo{author}{Wabersich, K.P.},
  \bibinfo{author}{Menner, M.}, \bibinfo{author}{Zeilinger, M.N.},
  \bibinfo{year}{2020}.
\newblock \bibinfo{title}{Learning-based model predictive control: Toward safe
  learning in control}.
\newblock \bibinfo{journal}{Annual Review of Control, Robotics, and Autonomous
  Systems} \bibinfo{volume}{3}, \bibinfo{pages}{269--296}.
\bibitem[{Hoang and Kang(2019)}]{Hoang2019}
\bibinfo{author}{Hoang, D.T.}, \bibinfo{author}{Kang, H.J.},
  \bibinfo{year}{2019}.
\newblock \bibinfo{title}{Rolling element bearing fault diagnosis using
  convolutional neural network and vibration image}.
\newblock \bibinfo{journal}{Cognitive Systems Research} \bibinfo{volume}{53},
  \bibinfo{pages}{42--50}.
\newblock \bibinfo{note}{Advanced Intelligent Computing}.
\bibitem[{Hoskins and Himmelblau(1992)}]{hoskins1992process}
\bibinfo{author}{Hoskins, J.}, \bibinfo{author}{Himmelblau, D.},
  \bibinfo{year}{1992}.
\newblock \bibinfo{title}{Process control via artificial neural networks and
  reinforcement learning}.
\newblock \bibinfo{journal}{Computers \& Chemical Engineering}
  \bibinfo{volume}{16}, \bibinfo{pages}{241--251}.
\bibitem[{Huang et~al.(1997)Huang, Shah and Kwok}]{huang1997good}
\bibinfo{author}{Huang, B.}, \bibinfo{author}{Shah, S.L.},
  \bibinfo{author}{Kwok, E.}, \bibinfo{year}{1997}.
\newblock \bibinfo{title}{Good, bad or optimal? {P}erformance assessment of
  multivariable processes}.
\newblock \bibinfo{journal}{Automatica} \bibinfo{volume}{33},
  \bibinfo{pages}{1175--1183}.
\bibitem[{Huang and Von~Lilienfeld(2016)}]{huang2016communication}
\bibinfo{author}{Huang, B.}, \bibinfo{author}{Von~Lilienfeld, O.A.},
  \bibinfo{year}{2016}.
\newblock \bibinfo{title}{Communication: Understanding molecular
  representations in machine learning: The role of uniqueness and target
  similarity}.
\bibitem[{Huang et~al.(2010)Huang, Biegler and Patwardhan}]{huang2010fast}
\bibinfo{author}{Huang, R.}, \bibinfo{author}{Biegler, L.T.},
  \bibinfo{author}{Patwardhan, S.C.}, \bibinfo{year}{2010}.
\newblock \bibinfo{title}{Fast offset-free nonlinear model predictive control
  based on moving horizon estimation}.
\newblock \bibinfo{journal}{Industrial \& Engineering Chemistry Research}
  \bibinfo{volume}{49}, \bibinfo{pages}{7882--7890}.
\bibitem[{Hussain(1999)}]{hussain1999review}
\bibinfo{author}{Hussain, M.A.}, \bibinfo{year}{1999}.
\newblock \bibinfo{title}{Review of the applications of neural networks in
  chemical process control—simulation and online implementation}.
\newblock \bibinfo{journal}{Artificial intelligence in engineering}
  \bibinfo{volume}{13}, \bibinfo{pages}{55--68}.
\bibitem[{Huster et~al.(2019)Huster, Schweidtmann and
  Mitsos}]{huster2019working}
\bibinfo{author}{Huster, W.R.}, \bibinfo{author}{Schweidtmann, A.M.},
  \bibinfo{author}{Mitsos, A.}, \bibinfo{year}{2019}.
\newblock \bibinfo{title}{Working fluid selection for organic rankine cycles
  via deterministic global optimization of design and operation}.
\newblock \bibinfo{journal}{Optimization and Engineering} ,
  \bibinfo{pages}{1--20}.
\bibitem[{Ibrahim et~al.(2018)Ibrahim, Jobson, Li and
  Guill{\'e}n-Gos{\'a}lbez}]{ibrahim2018optimization}
\bibinfo{author}{Ibrahim, D.}, \bibinfo{author}{Jobson, M.},
  \bibinfo{author}{Li, J.}, \bibinfo{author}{Guill{\'e}n-Gos{\'a}lbez, G.},
  \bibinfo{year}{2018}.
\newblock \bibinfo{title}{Optimization-based design of crude oil distillation
  units using surrogate column models and a support vector machine}.
\newblock \bibinfo{journal}{Chemical Engineering Research \& Design}
  \bibinfo{volume}{134}, \bibinfo{pages}{212--225}.
\bibitem[{Imtiaz and Shah(2008)}]{Imtiaz2008}
\bibinfo{author}{Imtiaz, S.A.}, \bibinfo{author}{Shah, S.L.},
  \bibinfo{year}{2008}.
\newblock \bibinfo{title}{Treatment of missing values in process data
  analysis}.
\newblock \bibinfo{journal}{Canadian Journal of Chemical Engineering}
  \bibinfo{volume}{86}, \bibinfo{pages}{838--858}.
\bibitem[{Isaksson et~al.(2018)Isaksson, Harjunkoski and
  Sand}]{isaksson2018impact}
\bibinfo{author}{Isaksson, A.J.}, \bibinfo{author}{Harjunkoski, I.},
  \bibinfo{author}{Sand, G.}, \bibinfo{year}{2018}.
\newblock \bibinfo{title}{The impact of digitalization on the future of control
  and operations}.
\newblock \bibinfo{journal}{Computers \& Chemical Engineering}
  \bibinfo{volume}{114}, \bibinfo{pages}{122--129}.
\bibitem[{Jain et~al.(2018)Jain, Chakraborty, Pistikopoulos and
  Mannan}]{Jain2018}
\bibinfo{author}{Jain, P.}, \bibinfo{author}{Chakraborty, A.},
  \bibinfo{author}{Pistikopoulos, E.N.}, \bibinfo{author}{Mannan, M.S.},
  \bibinfo{year}{2018}.
\newblock \bibinfo{title}{Resilience-based process upset event prediction
  analysis for uncertainty management using {Bayesian} deep learning:
  Application to a polyvinyl chloride process system}.
\newblock \bibinfo{journal}{Industrial \& Engineering Chemistry Research}
  \bibinfo{volume}{57}, \bibinfo{pages}{14822--14836}.
\bibitem[{Janet and Kulik(2017)}]{janet2017resolving}
\bibinfo{author}{Janet, J.P.}, \bibinfo{author}{Kulik, H.J.},
  \bibinfo{year}{2017}.
\newblock \bibinfo{title}{Resolving transition metal chemical space: Feature
  selection for machine learning and structure--property relationships}.
\newblock \bibinfo{journal}{The Journal of Physical Chemistry A}
  \bibinfo{volume}{121}, \bibinfo{pages}{8939--8954}.
\bibitem[{Jardine et~al.(2006)Jardine, Lin and Banjevic}]{Jardine2006}
\bibinfo{author}{Jardine, A.K.}, \bibinfo{author}{Lin, D.},
  \bibinfo{author}{Banjevic, D.}, \bibinfo{year}{2006}.
\newblock \bibinfo{title}{A review on machinery diagnostics and prognostics
  implementing condition-based maintenance}.
\newblock \bibinfo{journal}{Mechanical Systems and Signal Processing}
  \bibinfo{volume}{20}, \bibinfo{pages}{1483--1510}.
\bibitem[{Jia et~al.(2009)Jia, Davis, Muzzio and
  Ierapetritou}]{jia2009predictive}
\bibinfo{author}{Jia, Z.}, \bibinfo{author}{Davis, E.},
  \bibinfo{author}{Muzzio, F.J.}, \bibinfo{author}{Ierapetritou, M.G.},
  \bibinfo{year}{2009}.
\newblock \bibinfo{title}{Predictive modeling for pharmaceutical processes
  using {Kriging} and response surface}.
\newblock \bibinfo{journal}{Journal of Pharmaceutical Innovation}
  \bibinfo{volume}{4}, \bibinfo{pages}{174--186}.
\bibitem[{Jiang et~al.(2019)Jiang, Yan and Huang}]{jiang2019review}
\bibinfo{author}{Jiang, Q.}, \bibinfo{author}{Yan, X.}, \bibinfo{author}{Huang,
  B.}, \bibinfo{year}{2019}.
\newblock \bibinfo{title}{Review and perspectives of data-driven distributed
  monitoring for industrial plant-wide processes}.
\newblock \bibinfo{journal}{Industrial \& Engineering Chemistry Research}
  \bibinfo{volume}{58}, \bibinfo{pages}{12899--12912}.
\bibitem[{Johansen(2011)}]{johansen2011introduction}
\bibinfo{author}{Johansen, T.A.}, \bibinfo{year}{2011}.
\newblock \bibinfo{title}{Introduction to nonlinear model predictive control
  and moving horizon estimation}.
\newblock \bibinfo{journal}{Selected topics on constrained and nonlinear
  control} \bibinfo{volume}{1}, \bibinfo{pages}{1--53}.
\bibitem[{Kadlec et~al.(2011)Kadlec, Grbić and Gabrys}]{Kadlec2011}
\bibinfo{author}{Kadlec, P.}, \bibinfo{author}{Grbić, R.},
  \bibinfo{author}{Gabrys, B.}, \bibinfo{year}{2011}.
\newblock \bibinfo{title}{Review of adaptation mechanisms for data-driven soft
  sensors}.
\newblock \bibinfo{journal}{Computers \& Chemical Engineering}
  \bibinfo{volume}{35}, \bibinfo{pages}{1--24}.
\bibitem[{Karpatne et~al.(2017)Karpatne, Watkins, Read and
  Kumar}]{karpatne2017physics}
\bibinfo{author}{Karpatne, A.}, \bibinfo{author}{Watkins, W.},
  \bibinfo{author}{Read, J.}, \bibinfo{author}{Kumar, V.},
  \bibinfo{year}{2017}.
\newblock \bibinfo{title}{Physics-guided neural networks (pgnn): An application
  in lake temperature modeling}.
\newblock \bibinfo{journal}{arXiv preprint arXiv:1710.11431} .
\bibitem[{Khan and Madden(2009)}]{khan2009survey}
\bibinfo{author}{Khan, S.S.}, \bibinfo{author}{Madden, M.G.},
  \bibinfo{year}{2009}.
\newblock \bibinfo{title}{A survey of recent trends in one class
  classification}, in: \bibinfo{booktitle}{Irish conference on artificial
  intelligence and cognitive science}, \bibinfo{organization}{Springer}. pp.
  \bibinfo{pages}{188--197}.
\bibitem[{Lasi et~al.(2014)Lasi, Fettke, Kemper, Feld and
  Hoffmann}]{lasi2014industry}
\bibinfo{author}{Lasi, H.}, \bibinfo{author}{Fettke, P.},
  \bibinfo{author}{Kemper, H.G.}, \bibinfo{author}{Feld, T.},
  \bibinfo{author}{Hoffmann, M.}, \bibinfo{year}{2014}.
\newblock \bibinfo{title}{Industry 4.0}.
\newblock \bibinfo{journal}{Business \& information systems engineering}
  \bibinfo{volume}{6}, \bibinfo{pages}{239--242}.
\bibitem[{Laur{\'\i} et~al.(2010)Laur{\'\i}, Rossiter, Sanchis and
  Mart{\'\i}nez}]{lauri2010data}
\bibinfo{author}{Laur{\'\i}, D.}, \bibinfo{author}{Rossiter, J.A.},
  \bibinfo{author}{Sanchis, J.}, \bibinfo{author}{Mart{\'\i}nez, M.},
  \bibinfo{year}{2010}.
\newblock \bibinfo{title}{Data-driven latent-variable model-based predictive
  control for continuous processes}.
\newblock \bibinfo{journal}{Journal of Process Control} \bibinfo{volume}{20},
  \bibinfo{pages}{1207--1219}.
\bibitem[{Lee et~al.(2011)Lee, Kim, Byeon, Sung and Edgar}]{lee2011relay}
\bibinfo{author}{Lee, J.}, \bibinfo{author}{Kim, J.S.}, \bibinfo{author}{Byeon,
  J.}, \bibinfo{author}{Sung, S.W.}, \bibinfo{author}{Edgar, T.F.},
  \bibinfo{year}{2011}.
\newblock \bibinfo{title}{Relay feedback identification for processes under
  drift and noisy environments}.
\newblock \bibinfo{journal}{AIChE journal} \bibinfo{volume}{57},
  \bibinfo{pages}{1809--1816}.
\bibitem[{Lee et~al.(2018)Lee, Shin and Realff}]{Lee2018}
\bibinfo{author}{Lee, J.H.}, \bibinfo{author}{Shin, J.},
  \bibinfo{author}{Realff, M.J.}, \bibinfo{year}{2018}.
\newblock \bibinfo{title}{Machine learning: Overview of the recent progresses
  and implications for the process systems engineering field}.
\newblock \bibinfo{journal}{Computers \& Chemical Engineering}
  \bibinfo{volume}{114}, \bibinfo{pages}{111--121}.
\bibitem[{Lin and Tsai(2020)}]{Lin2020}
\bibinfo{author}{Lin, W.C.}, \bibinfo{author}{Tsai, C.F.},
  \bibinfo{year}{2020}.
\newblock \bibinfo{title}{Missing value imputation: a review and analysis of
  the literature (2006–2017)}.
\newblock \bibinfo{journal}{Artificial Intelligence Review}
  \bibinfo{volume}{53}, \bibinfo{pages}{1487--1509}.
\bibitem[{Liu et~al.(2020)Liu, Yang, Li, Xu and Deng}]{Liu2020}
\bibinfo{author}{Liu, C.}, \bibinfo{author}{Yang, S.X.}, \bibinfo{author}{Li,
  X.}, \bibinfo{author}{Xu, L.}, \bibinfo{author}{Deng, L.},
  \bibinfo{year}{2020}.
\newblock \bibinfo{title}{Noise level penalizing robust gaussian process
  regression for {NIR} spectroscopy quantitative analysis}.
\newblock \bibinfo{journal}{Chemometrics and Intelligent Laboratory Systems}
  \bibinfo{volume}{201}.
\bibitem[{Lizotte(2008)}]{lizotte2008practical}
\bibinfo{author}{Lizotte, D.J.}, \bibinfo{year}{2008}.
\newblock \bibinfo{title}{Practical {Bayesian} optimization}.
\newblock \bibinfo{publisher}{University of Alberta}.
\bibitem[{Ljung and Gunnarsson(1990)}]{ljung1990adaptation}
\bibinfo{author}{Ljung, L.}, \bibinfo{author}{Gunnarsson, S.},
  \bibinfo{year}{1990}.
\newblock \bibinfo{title}{Adaptation and tracking in system identification—a
  survey}.
\newblock \bibinfo{journal}{Automatica} \bibinfo{volume}{26},
  \bibinfo{pages}{7--21}.
\bibitem[{Louizos et~al.(2018)Louizos, Welling and
  Kingma}]{louizos2017learning}
\bibinfo{author}{Louizos, C.}, \bibinfo{author}{Welling, M.},
  \bibinfo{author}{Kingma, D.P.}, \bibinfo{year}{2018}.
\newblock \bibinfo{title}{Learning sparse neural networks through $l_0$
  regularization}, in: \bibinfo{booktitle}{International Conference on Learning
  Representations}.
\bibitem[{Lovelett et~al.(2020)Lovelett, Dietrich, Lee and
  Kevrekidis}]{lovelett2020some}
\bibinfo{author}{Lovelett, R.J.}, \bibinfo{author}{Dietrich, F.},
  \bibinfo{author}{Lee, S.}, \bibinfo{author}{Kevrekidis, I.G.},
  \bibinfo{year}{2020}.
\newblock \bibinfo{title}{Some manifold learning considerations toward explicit
  model predictive control}.
\newblock \bibinfo{journal}{AIChE Journal} \bibinfo{volume}{66},
  \bibinfo{pages}{e16881}.
\bibitem[{MacGregor and Cinar(2012)}]{macgregor2012monitoring}
\bibinfo{author}{MacGregor, J.}, \bibinfo{author}{Cinar, A.},
  \bibinfo{year}{2012}.
\newblock \bibinfo{title}{Monitoring, fault diagnosis, fault-tolerant control
  and optimization: Data driven methods}.
\newblock \bibinfo{journal}{Computers \& Chemical Engineering}
  \bibinfo{volume}{47}, \bibinfo{pages}{111--120}.
\bibitem[{Manng{\aa}rd et~al.(2018)Manng{\aa}rd, Kronqvist and
  B{\"o}ling}]{manngaard2018structural}
\bibinfo{author}{Manng{\aa}rd, M.}, \bibinfo{author}{Kronqvist, J.},
  \bibinfo{author}{B{\"o}ling, J.M.}, \bibinfo{year}{2018}.
\newblock \bibinfo{title}{Structural learning in artificial neural networks
  using sparse optimization}.
\newblock \bibinfo{journal}{Neurocomputing} \bibinfo{volume}{272},
  \bibinfo{pages}{660--667}.
\bibitem[{Martelli and Amaldi(2014)}]{martelli2014pgs}
\bibinfo{author}{Martelli, E.}, \bibinfo{author}{Amaldi, E.},
  \bibinfo{year}{2014}.
\newblock \bibinfo{title}{Pgs-com: a hybrid method for constrained non-smooth
  black-box optimization problems: brief review, novel algorithm and
  comparative evaluation}.
\newblock \bibinfo{journal}{Computers \& Chemical Engineering}
  \bibinfo{volume}{63}, \bibinfo{pages}{108--139}.
\bibitem[{McBride and Sundmacher(2019)}]{mcbride2019overview}
\bibinfo{author}{McBride, K.}, \bibinfo{author}{Sundmacher, K.},
  \bibinfo{year}{2019}.
\newblock \bibinfo{title}{Overview of surrogate modeling in chemical process
  engineering}.
\newblock \bibinfo{journal}{Chemie Ingenieur Technik} \bibinfo{volume}{91},
  \bibinfo{pages}{228--239}.
\bibitem[{Medsker(2012)}]{medsker2012hybrid}
\bibinfo{author}{Medsker, L.R.}, \bibinfo{year}{2012}.
\newblock \bibinfo{title}{Hybrid neural network and expert systems}.
\newblock \bibinfo{publisher}{Springer Science \& Business Media}.
\bibitem[{Mesbah(2018)}]{mesbah2018stochastic}
\bibinfo{author}{Mesbah, A.}, \bibinfo{year}{2018}.
\newblock \bibinfo{title}{Stochastic model predictive control with active
  uncertainty learning: A survey on dual control}.
\newblock \bibinfo{journal}{Annual Reviews in Control} \bibinfo{volume}{45},
  \bibinfo{pages}{107--117}.
\bibitem[{Mesquita et~al.(2019)Mesquita, Gomes, Corona, Souza and
  Nobre}]{Mesquita2019}
\bibinfo{author}{Mesquita, D.P.}, \bibinfo{author}{Gomes, J.P.},
  \bibinfo{author}{Corona, F.}, \bibinfo{author}{Souza, A.H.},
  \bibinfo{author}{Nobre, J.S.}, \bibinfo{year}{2019}.
\newblock \bibinfo{title}{Gaussian kernels for incomplete data}.
\newblock \bibinfo{journal}{Applied Soft Computing Journal}
  \bibinfo{volume}{77}, \bibinfo{pages}{356--365}.
\bibitem[{Metta et~al.(2020)Metta, Ramachandran and
  Ierapetritou}]{metta2020novel}
\bibinfo{author}{Metta, N.}, \bibinfo{author}{Ramachandran, R.},
  \bibinfo{author}{Ierapetritou, M.}, \bibinfo{year}{2020}.
\newblock \bibinfo{title}{A novel adaptive sampling based methodology for
  feasible region identification of compute intensive models using artificial
  neural network}.
\newblock \bibinfo{journal}{AIChE Journal} , \bibinfo{pages}{e17095}.
\bibitem[{Misener and Floudas(2014)}]{misener2014antigone}
\bibinfo{author}{Misener, R.}, \bibinfo{author}{Floudas, C.A.},
  \bibinfo{year}{2014}.
\newblock \bibinfo{title}{Antigone: algorithms for continuous/integer global
  optimization of nonlinear equations}.
\newblock \bibinfo{journal}{Journal of Global Optimization}
  \bibinfo{volume}{59}, \bibinfo{pages}{503--526}.
\bibitem[{Mi{\v{s}}i{\'c}(2020)}]{mivsic2020optimization}
\bibinfo{author}{Mi{\v{s}}i{\'c}, V.V.}, \bibinfo{year}{2020}.
\newblock \bibinfo{title}{Optimization of tree ensembles}.
\newblock \bibinfo{journal}{Operations Research} \bibinfo{volume}{68},
  \bibinfo{pages}{1605--1624}.
\bibitem[{Mistry et~al.(2020)Mistry, Letsios, Krennrich, Lee and
  Misener}]{mistry2020mixed}
\bibinfo{author}{Mistry, M.}, \bibinfo{author}{Letsios, D.},
  \bibinfo{author}{Krennrich, G.}, \bibinfo{author}{Lee, R.M.},
  \bibinfo{author}{Misener, R.}, \bibinfo{year}{2020}.
\newblock \bibinfo{title}{Mixed-integer convex nonlinear optimization with
  gradient-boosted trees embedded}.
\newblock \bibinfo{journal}{INFORMS Journal on Computing} .
\bibitem[{Montgomery et~al.(1994)Montgomery, Keats, Runger and
  Messina}]{montgomery1994integrating}
\bibinfo{author}{Montgomery, D.C.}, \bibinfo{author}{Keats, J.B.},
  \bibinfo{author}{Runger, G.C.}, \bibinfo{author}{Messina, W.S.},
  \bibinfo{year}{1994}.
\newblock \bibinfo{title}{Integrating statistical process control and
  engineering process control}.
\newblock \bibinfo{journal}{Journal of quality Technology}
  \bibinfo{volume}{26}, \bibinfo{pages}{79--87}.
\bibitem[{Nannicini and Belotti(2012)}]{nannicini2012rounding}
\bibinfo{author}{Nannicini, G.}, \bibinfo{author}{Belotti, P.},
  \bibinfo{year}{2012}.
\newblock \bibinfo{title}{Rounding-based heuristics for nonconvex {MINLP}s}.
\newblock \bibinfo{journal}{Mathematical Programming Computation}
  \bibinfo{volume}{4}, \bibinfo{pages}{1--31}.
\bibitem[{Naz\'abal et~al.(2020)Naz\'abal, Olmos, Ghahramani and
  Valera}]{Nazabal2020}
\bibinfo{author}{Naz\'abal, A.}, \bibinfo{author}{Olmos, P.M.},
  \bibinfo{author}{Ghahramani, Z.}, \bibinfo{author}{Valera, I.},
  \bibinfo{year}{2020}.
\newblock \bibinfo{title}{Handling incomplete heterogeneous data using vaes}.
\newblock \bibinfo{journal}{Pattern Recognition} \bibinfo{volume}{107}.
\bibitem[{Nguyen et~al.(2018)Nguyen, Fouladirad and Grall}]{Nguyen2018}
\bibinfo{author}{Nguyen, K.T.}, \bibinfo{author}{Fouladirad, M.},
  \bibinfo{author}{Grall, A.}, \bibinfo{year}{2018}.
\newblock \bibinfo{title}{Model selection for degradation modeling and
  prognosis with health monitoring data}.
\newblock \bibinfo{journal}{Reliability Engineering \& System Safety}
  \bibinfo{volume}{169}, \bibinfo{pages}{105--116}.
\bibitem[{Ning and You(2018)}]{ning2018ddsro}
\bibinfo{author}{Ning, C.}, \bibinfo{author}{You, F.}, \bibinfo{year}{2018}.
\newblock \bibinfo{title}{Data-driven stochastic robust optimization: General
  computational framework and algorithm leveraging machine learning for
  optimization under uncertainty in the big data era}.
\newblock \bibinfo{journal}{Computers \& Chemical Engineering}
  \bibinfo{volume}{111}, \bibinfo{pages}{115--133}.
\bibitem[{Ning and You(2019)}]{ning2019optimization}
\bibinfo{author}{Ning, C.}, \bibinfo{author}{You, F.}, \bibinfo{year}{2019}.
\newblock \bibinfo{title}{Optimization under uncertainty in the era of big data
  and deep learning: When machine learning meets mathematical programming}.
\newblock \bibinfo{journal}{Computers \& Chemical Engineering}
  \bibinfo{volume}{125}, \bibinfo{pages}{434--448}.
\bibitem[{Olofsson et~al.(2018)Olofsson, Mehrian, Calandra, Geris, Deisenroth
  and Misener}]{olofsson2018bayesian}
\bibinfo{author}{Olofsson, S.}, \bibinfo{author}{Mehrian, M.},
  \bibinfo{author}{Calandra, R.}, \bibinfo{author}{Geris, L.},
  \bibinfo{author}{Deisenroth, M.P.}, \bibinfo{author}{Misener, R.},
  \bibinfo{year}{2018}.
\newblock \bibinfo{title}{Bayesian multiobjective optimisation with mixed
  analytical and black-box functions: Application to tissue engineering}.
\newblock \bibinfo{journal}{IEEE Transactions on Biomedical Engineering}
  \bibinfo{volume}{66}, \bibinfo{pages}{727--739}.
\bibitem[{Palmer and Realff(2002)}]{palmer2002metamodeling}
\bibinfo{author}{Palmer, K.}, \bibinfo{author}{Realff, M.},
  \bibinfo{year}{2002}.
\newblock \bibinfo{title}{Metamodeling approach to optimization of steady-state
  flowsheet simulations: Model generation}.
\newblock \bibinfo{journal}{Chemical Engineering Research \& Design}
  \bibinfo{volume}{80}, \bibinfo{pages}{760--772}.
\bibitem[{Pamula et~al.(2011)Pamula, Deka and Nandi}]{Pamula2011}
\bibinfo{author}{Pamula, R.}, \bibinfo{author}{Deka, J.K.},
  \bibinfo{author}{Nandi, S.}, \bibinfo{year}{2011}.
\newblock \bibinfo{title}{An outlier detection method based on clustering}, pp.
  \bibinfo{pages}{253--256}.
\bibitem[{Pan and Yang(2009)}]{pan2009survey}
\bibinfo{author}{Pan, S.J.}, \bibinfo{author}{Yang, Q.}, \bibinfo{year}{2009}.
\newblock \bibinfo{title}{A survey on transfer learning}.
\newblock \bibinfo{journal}{IEEE Transactions on knowledge and data
  engineering} \bibinfo{volume}{22}, \bibinfo{pages}{1345--1359}.
\bibitem[{Pattison et~al.(2016)Pattison, Touretzky, Johansson, Harjunkoski and
  Baldea}]{pattison2016optimal}
\bibinfo{author}{Pattison, R.C.}, \bibinfo{author}{Touretzky, C.R.},
  \bibinfo{author}{Johansson, T.}, \bibinfo{author}{Harjunkoski, I.},
  \bibinfo{author}{Baldea, M.}, \bibinfo{year}{2016}.
\newblock \bibinfo{title}{Optimal process operations in fast-changing
  electricity markets: framework for scheduling with low-order dynamic models
  and an air separation application}.
\newblock \bibinfo{journal}{Industrial \& Engineering Chemistry Research}
  \bibinfo{volume}{55}, \bibinfo{pages}{4562--4584}.
\bibitem[{Paulson and Mesbah(2020)}]{paulson2020approximate}
\bibinfo{author}{Paulson, J.A.}, \bibinfo{author}{Mesbah, A.},
  \bibinfo{year}{2020}.
\newblock \bibinfo{title}{Approximate closed-loop robust model predictive
  control with guaranteed stability and constraint satisfaction}.
\newblock \bibinfo{journal}{IEEE Control Systems Letters} \bibinfo{volume}{4},
  \bibinfo{pages}{719--724}.
\bibitem[{Pham et~al.(2020)Pham, Kim and Kim}]{Pham2020}
\bibinfo{author}{Pham, M.T.}, \bibinfo{author}{Kim, J.M.},
  \bibinfo{author}{Kim, C.H.}, \bibinfo{year}{2020}.
\newblock \bibinfo{title}{Accurate bearing fault diagnosis under variable shaft
  speed using convolutional neural networks and vibration spectrogram}.
\newblock \bibinfo{journal}{Applied Sciences (Switzerland)}
  \bibinfo{volume}{10}.
\bibitem[{Pich{\'e} et~al.(2001)Pich{\'e}, Larachi and
  Grandjean}]{piche2001flooding}
\bibinfo{author}{Pich{\'e}, S.}, \bibinfo{author}{Larachi, F.},
  \bibinfo{author}{Grandjean, B.P.}, \bibinfo{year}{2001}.
\newblock \bibinfo{title}{Flooding capacity in packed towers: database,
  correlations, and analysis}.
\newblock \bibinfo{journal}{Industrial \& Engineering Chemistry Research}
  \bibinfo{volume}{40}, \bibinfo{pages}{476--487}.
\bibitem[{Pistikopoulos et~al.(2021)Pistikopoulos, Barbosa-Povoa, Lee, Misener,
  Mitsos, Reklaitis, Venkatasubramanian, You and
  Gani}]{PISTIKOPOULOS2021107252}
\bibinfo{author}{Pistikopoulos, E.N.}, \bibinfo{author}{Barbosa-Povoa, A.},
  \bibinfo{author}{Lee, J.H.}, \bibinfo{author}{Misener, R.},
  \bibinfo{author}{Mitsos, A.}, \bibinfo{author}{Reklaitis, G.V.},
  \bibinfo{author}{Venkatasubramanian, V.}, \bibinfo{author}{You, F.},
  \bibinfo{author}{Gani, R.}, \bibinfo{year}{2021}.
\newblock \bibinfo{title}{Process systems engineering – the generation next?}
\newblock \bibinfo{journal}{Computers \& Chemical Engineering}
  \bibinfo{volume}{147}, \bibinfo{pages}{107252}.
\newblock \URLprefix
  \url{https://www.sciencedirect.com/science/article/pii/S0098135421000302},
  \DOIprefix\doi{https://doi.org/10.1016/j.compchemeng.2021.107252}.
\bibitem[{Ponzoni et~al.(2017)Ponzoni, Sebasti{\'a}n-P{\'e}rez,
  Requena-Triguero, Roca, Mart{\'\i}nez, Cravero, D{\'\i}az, P{\'a}ez,
  Array{\'a}s, Adrio et~al.}]{ponzoni2017hybridizing}
\bibinfo{author}{Ponzoni, I.}, \bibinfo{author}{Sebasti{\'a}n-P{\'e}rez, V.},
  \bibinfo{author}{Requena-Triguero, C.}, \bibinfo{author}{Roca, C.},
  \bibinfo{author}{Mart{\'\i}nez, M.J.}, \bibinfo{author}{Cravero, F.},
  \bibinfo{author}{D{\'\i}az, M.F.}, \bibinfo{author}{P{\'a}ez, J.A.},
  \bibinfo{author}{Array{\'a}s, R.G.}, \bibinfo{author}{Adrio, J.}, et~al.,
  \bibinfo{year}{2017}.
\newblock \bibinfo{title}{Hybridizing feature selection and feature learning
  approaches in {QSAR} modeling for drug discovery}.
\newblock \bibinfo{journal}{Scientific reports} \bibinfo{volume}{7},
  \bibinfo{pages}{1--19}.
\bibitem[{Psichogios and Ungar(1992)}]{psichogios1992hybrid}
\bibinfo{author}{Psichogios, D.C.}, \bibinfo{author}{Ungar, L.H.},
  \bibinfo{year}{1992}.
\newblock \bibinfo{title}{A hybrid neural network-first principles approach to
  process modeling}.
\newblock \bibinfo{journal}{AIChE Journal} \bibinfo{volume}{38},
  \bibinfo{pages}{1499--1511}.
\bibitem[{Qin(2012)}]{qin2012survey}
\bibinfo{author}{Qin, S.J.}, \bibinfo{year}{2012}.
\newblock \bibinfo{title}{Survey on data-driven industrial process monitoring
  and diagnosis}.
\newblock \bibinfo{journal}{Annual reviews in control} \bibinfo{volume}{36},
  \bibinfo{pages}{220--234}.
\bibitem[{Qin and Chiang(2019)}]{qin2019advances}
\bibinfo{author}{Qin, S.J.}, \bibinfo{author}{Chiang, L.H.},
  \bibinfo{year}{2019}.
\newblock \bibinfo{title}{Advances and opportunities in machine learning for
  process data analytics}.
\newblock \bibinfo{journal}{Computers \& Chemical Engineering}
  \bibinfo{volume}{126}, \bibinfo{pages}{465--473}.
\bibitem[{Qin et~al.(2020)Qin, Dong, Zhu, Wang and Liu}]{qin2020bridging}
\bibinfo{author}{Qin, S.J.}, \bibinfo{author}{Dong, Y.}, \bibinfo{author}{Zhu,
  Q.}, \bibinfo{author}{Wang, J.}, \bibinfo{author}{Liu, Q.},
  \bibinfo{year}{2020}.
\newblock \bibinfo{title}{Bridging systems theory and data science: A unifying
  review of dynamic latent variable analytics and process monitoring}.
\newblock \bibinfo{journal}{Annual Reviews in Control} .
\bibitem[{Qui{\~n}ones-Grueiro et~al.(2019)Qui{\~n}ones-Grueiro, Prieto-Moreno,
  Verde and Llanes-Santiago}]{quinones2019data}
\bibinfo{author}{Qui{\~n}ones-Grueiro, M.}, \bibinfo{author}{Prieto-Moreno,
  A.}, \bibinfo{author}{Verde, C.}, \bibinfo{author}{Llanes-Santiago, O.},
  \bibinfo{year}{2019}.
\newblock \bibinfo{title}{Data-driven monitoring of multimode continuous
  processes: A review}.
\newblock \bibinfo{journal}{Chemometrics and Intelligent Laboratory Systems}
  \bibinfo{volume}{189}, \bibinfo{pages}{56--71}.
\bibitem[{Quirante et~al.(2015)Quirante, Javaloyes, Ruiz-Femenia and
  Caballero}]{quirante2015optimization}
\bibinfo{author}{Quirante, N.}, \bibinfo{author}{Javaloyes, J.},
  \bibinfo{author}{Ruiz-Femenia, R.}, \bibinfo{author}{Caballero, J.A.},
  \bibinfo{year}{2015}.
\newblock \bibinfo{title}{Optimization of chemical processes using surrogate
  models based on a {Kriging} interpolation}, in: \bibinfo{booktitle}{Computer
  Aided Chemical Engineering}. \bibinfo{publisher}{Elsevier}.
  volume~\bibinfo{volume}{37}, pp. \bibinfo{pages}{179--184}.
\bibitem[{Raissi(2018)}]{raissi2018deep}
\bibinfo{author}{Raissi, M.}, \bibinfo{year}{2018}.
\newblock \bibinfo{title}{Deep hidden physics models: Deep learning of
  nonlinear partial differential equations}.
\newblock \bibinfo{journal}{The Journal of Machine Learning Research}
  \bibinfo{volume}{19}, \bibinfo{pages}{932--955}.
\bibitem[{Raissi et~al.(2019)Raissi, Perdikaris and
  Karniadakis}]{raissi2019physics}
\bibinfo{author}{Raissi, M.}, \bibinfo{author}{Perdikaris, P.},
  \bibinfo{author}{Karniadakis, G.E.}, \bibinfo{year}{2019}.
\newblock \bibinfo{title}{Physics-informed neural networks: A deep learning
  framework for solving forward and inverse problems involving nonlinear
  partial differential equations}.
\newblock \bibinfo{journal}{Journal of Computational Physics}
  \bibinfo{volume}{378}, \bibinfo{pages}{686--707}.
\bibitem[{Rall et~al.(2019)Rall, Menne, Schweidtmann, Kamp, von Kolzenberg,
  Mitsos and Wessling}]{rall2019rational}
\bibinfo{author}{Rall, D.}, \bibinfo{author}{Menne, D.},
  \bibinfo{author}{Schweidtmann, A.M.}, \bibinfo{author}{Kamp, J.},
  \bibinfo{author}{von Kolzenberg, L.}, \bibinfo{author}{Mitsos, A.},
  \bibinfo{author}{Wessling, M.}, \bibinfo{year}{2019}.
\newblock \bibinfo{title}{Rational design of ion separation membranes}.
\newblock \bibinfo{journal}{Journal of Membrane Science} \bibinfo{volume}{569},
  \bibinfo{pages}{209--219}.
\bibitem[{Rall et~al.(2020)Rall, Schweidtmann, Aumeier, Kamp, Karwe, Ostendorf,
  Mitsos and Wessling}]{rall2020simultaneous}
\bibinfo{author}{Rall, D.}, \bibinfo{author}{Schweidtmann, A.M.},
  \bibinfo{author}{Aumeier, B.M.}, \bibinfo{author}{Kamp, J.},
  \bibinfo{author}{Karwe, J.}, \bibinfo{author}{Ostendorf, K.},
  \bibinfo{author}{Mitsos, A.}, \bibinfo{author}{Wessling, M.},
  \bibinfo{year}{2020}.
\newblock \bibinfo{title}{Simultaneous rational design of ion separation
  membranes and processes}.
\newblock \bibinfo{journal}{Journal of Membrane Science} \bibinfo{volume}{600},
  \bibinfo{pages}{117860}.
\bibitem[{Rasmussen(2003)}]{rasmussen2003gaussian}
\bibinfo{author}{Rasmussen, C.E.}, \bibinfo{year}{2003}.
\newblock \bibinfo{title}{Gaussian processes in machine learning}, in:
  \bibinfo{booktitle}{Summer school on machine learning},
  \bibinfo{organization}{Springer}. pp. \bibinfo{pages}{63--71}.
\bibitem[{Rawlings and Bakshi(2006)}]{rawlings2006particle}
\bibinfo{author}{Rawlings, J.B.}, \bibinfo{author}{Bakshi, B.R.},
  \bibinfo{year}{2006}.
\newblock \bibinfo{title}{Particle filtering and moving horizon estimation}.
\newblock \bibinfo{journal}{Computers \& Chemical Engineering}
  \bibinfo{volume}{30}, \bibinfo{pages}{1529--1541}.
\bibitem[{Rawlings and Maravelias(2019)}]{rawlings2019bringing}
\bibinfo{author}{Rawlings, J.B.}, \bibinfo{author}{Maravelias, C.T.},
  \bibinfo{year}{2019}.
\newblock \bibinfo{title}{Bringing new technologies and approaches to the
  operation and control of chemical process systems}.
\newblock \bibinfo{journal}{AIChE Journal} \bibinfo{volume}{65},
  \bibinfo{pages}{e16615}.
\bibitem[{Riese and Gr{\"u}newald(2020)}]{riese2020challenges}
\bibinfo{author}{Riese, J.}, \bibinfo{author}{Gr{\"u}newald, M.},
  \bibinfo{year}{2020}.
\newblock \bibinfo{title}{Challenges and opportunities to enhance flexibility
  in design and operation of chemical processes}.
\newblock \bibinfo{journal}{Chemie Ingenieur Technik} \bibinfo{volume}{92},
  \bibinfo{pages}{1887--1897}.
\bibitem[{Rios and Sahinidis(2013)}]{rios2013derivative}
\bibinfo{author}{Rios, L.M.}, \bibinfo{author}{Sahinidis, N.V.},
  \bibinfo{year}{2013}.
\newblock \bibinfo{title}{Derivative-free optimization: a review of algorithms
  and comparison of software implementations}.
\newblock \bibinfo{journal}{Journal of Global Optimization}
  \bibinfo{volume}{56}, \bibinfo{pages}{1247--1293}.
\bibitem[{Rogers and Ierapetritou(2015)}]{rogers2015feasibility}
\bibinfo{author}{Rogers, A.}, \bibinfo{author}{Ierapetritou, M.},
  \bibinfo{year}{2015}.
\newblock \bibinfo{title}{Feasibility and flexibility analysis of black-box
  processes part 1: Surrogate-based feasibility analysis}.
\newblock \bibinfo{journal}{Chemical Engineering Science}
  \bibinfo{volume}{137}, \bibinfo{pages}{986--1004}.
\bibitem[{Ruff et~al.(2018)Ruff, Vandermeulen, Goernitz, Deecke, Siddiqui,
  Binder, M{\"u}ller and Kloft}]{ruff2018deep}
\bibinfo{author}{Ruff, L.}, \bibinfo{author}{Vandermeulen, R.},
  \bibinfo{author}{Goernitz, N.}, \bibinfo{author}{Deecke, L.},
  \bibinfo{author}{Siddiqui, S.A.}, \bibinfo{author}{Binder, A.},
  \bibinfo{author}{M{\"u}ller, E.}, \bibinfo{author}{Kloft, M.},
  \bibinfo{year}{2018}.
\newblock \bibinfo{title}{Deep one-class classification}, in:
  \bibinfo{booktitle}{International Conference on Machine Learning}, pp.
  \bibinfo{pages}{4393--4402}.
\bibitem[{Schweidtmann et~al.(2018)Schweidtmann, Clayton, Holmes, Bradford,
  Bourne and Lapkin}]{schweidtmann2018machine}
\bibinfo{author}{Schweidtmann, A.M.}, \bibinfo{author}{Clayton, A.D.},
  \bibinfo{author}{Holmes, N.}, \bibinfo{author}{Bradford, E.},
  \bibinfo{author}{Bourne, R.A.}, \bibinfo{author}{Lapkin, A.A.},
  \bibinfo{year}{2018}.
\newblock \bibinfo{title}{Machine learning meets continuous flow chemistry:
  Automated optimization towards the {Pareto} front of multiple objectives}.
\newblock \bibinfo{journal}{Chemical Engineering Journal}
  \bibinfo{volume}{352}, \bibinfo{pages}{277--282}.
\bibitem[{Schweidtmann et~al.(2019)Schweidtmann, Huster, L{\"u}thje and
  Mitsos}]{schweidtmann2019deterministic}
\bibinfo{author}{Schweidtmann, A.M.}, \bibinfo{author}{Huster, W.R.},
  \bibinfo{author}{L{\"u}thje, J.T.}, \bibinfo{author}{Mitsos, A.},
  \bibinfo{year}{2019}.
\newblock \bibinfo{title}{Deterministic global process optimization: Accurate
  (single-species) properties via artificial neural networks}.
\newblock \bibinfo{journal}{Computers \& Chemical Engineering}
  \bibinfo{volume}{121}, \bibinfo{pages}{67--74}.
\bibitem[{Schweidtmann and Mitsos(2019)}]{schweidtmann2019deterministic2}
\bibinfo{author}{Schweidtmann, A.M.}, \bibinfo{author}{Mitsos, A.},
  \bibinfo{year}{2019}.
\newblock \bibinfo{title}{Deterministic global optimization with artificial
  neural networks embedded}.
\newblock \bibinfo{journal}{Journal of Optimization Theory and Applications}
  \bibinfo{volume}{180}, \bibinfo{pages}{925--948}.
\bibitem[{Severson et~al.(2017)Severson, Molaro and Braatz}]{Severson2017}
\bibinfo{author}{Severson, K.A.}, \bibinfo{author}{Molaro, M.C.},
  \bibinfo{author}{Braatz, R.D.}, \bibinfo{year}{2017}.
\newblock \bibinfo{title}{Principal component analysis of process datasets with
  missing values}.
\newblock \bibinfo{journal}{Processes} \bibinfo{volume}{5}.
\bibitem[{Shahriari et~al.(2015)Shahriari, Swersky, Wang, Adams and
  De~Freitas}]{shahriari2015taking}
\bibinfo{author}{Shahriari, B.}, \bibinfo{author}{Swersky, K.},
  \bibinfo{author}{Wang, Z.}, \bibinfo{author}{Adams, R.P.},
  \bibinfo{author}{De~Freitas, N.}, \bibinfo{year}{2015}.
\newblock \bibinfo{title}{Taking the human out of the loop: A review of
  {Bayesian} optimization}.
\newblock \bibinfo{journal}{Proceedings of the IEEE} \bibinfo{volume}{104},
  \bibinfo{pages}{148--175}.
\bibitem[{Shang et~al.(2017)Shang, Huang and You}]{shang2017svc}
\bibinfo{author}{Shang, C.}, \bibinfo{author}{Huang, X.}, \bibinfo{author}{You,
  F.}, \bibinfo{year}{2017}.
\newblock \bibinfo{title}{Data-driven robust optimization based on kernel
  learning}.
\newblock \bibinfo{journal}{Computers \& Chemical Engineering}
  \bibinfo{volume}{106}, \bibinfo{pages}{464--479}.
\bibitem[{Shang and You(2019)}]{shang2019data}
\bibinfo{author}{Shang, C.}, \bibinfo{author}{You, F.}, \bibinfo{year}{2019}.
\newblock \bibinfo{title}{Data analytics and machine learning for smart process
  manufacturing: recent advances and perspectives in the big data era}.
\newblock \bibinfo{journal}{Engineering} \bibinfo{volume}{5},
  \bibinfo{pages}{1010--1016}.
\bibitem[{Shin et~al.(2019)Shin, Badgwell, Liu and Lee}]{shin2019reinforcement}
\bibinfo{author}{Shin, J.}, \bibinfo{author}{Badgwell, T.A.},
  \bibinfo{author}{Liu, K.H.}, \bibinfo{author}{Lee, J.H.},
  \bibinfo{year}{2019}.
\newblock \bibinfo{title}{Reinforcement learning--overview of recent progress
  and implications for process control}.
\newblock \bibinfo{journal}{Computers \& Chemical Engineering}
  \bibinfo{volume}{127}, \bibinfo{pages}{282--294}.
\bibitem[{Si and Wang(2019)}]{Si2019}
\bibinfo{author}{Si, Y.}, \bibinfo{author}{Wang, Y.}, \bibinfo{year}{2019}.
\newblock \bibinfo{title}{Two-step dynamic slow feature analysis for dynamic
  process monitoring}, in: \bibinfo{booktitle}{1st International Conference on
  Industrial Artificial Intelligence}.
\bibitem[{Simkoff et~al.(2020)Simkoff, Lejarza, Kelley, Tsay and
  Baldea}]{simkoff2020process}
\bibinfo{author}{Simkoff, J.M.}, \bibinfo{author}{Lejarza, F.},
  \bibinfo{author}{Kelley, M.T.}, \bibinfo{author}{Tsay, C.},
  \bibinfo{author}{Baldea, M.}, \bibinfo{year}{2020}.
\newblock \bibinfo{title}{Process control and energy efficiency}.
\newblock \bibinfo{journal}{Annual Review of Chemical and Biomolecular
  Engineering} \bibinfo{volume}{11}, \bibinfo{pages}{423--445}.
\bibitem[{Spivey et~al.(2010)Spivey, Hedengren and
  Edgar}]{spivey2010constrained}
\bibinfo{author}{Spivey, B.J.}, \bibinfo{author}{Hedengren, J.D.},
  \bibinfo{author}{Edgar, T.F.}, \bibinfo{year}{2010}.
\newblock \bibinfo{title}{Constrained nonlinear estimation for industrial
  process fouling}.
\newblock \bibinfo{journal}{Industrial \& Engineering Chemistry Research}
  \bibinfo{volume}{49}, \bibinfo{pages}{7824--7831}.
\bibitem[{Springenberg et~al.(2016)Springenberg, Klein, Falkner and
  Hutter}]{springenberg2016bayesian}
\bibinfo{author}{Springenberg, J.T.}, \bibinfo{author}{Klein, A.},
  \bibinfo{author}{Falkner, S.}, \bibinfo{author}{Hutter, F.},
  \bibinfo{year}{2016}.
\newblock \bibinfo{title}{Bayesian optimization with robust {Bayesian} neural
  networks}, in: \bibinfo{booktitle}{Proceedings of the 30th International
  Conference on Neural Information Processing Systems}, pp.
  \bibinfo{pages}{4141--4149}.
\bibitem[{Sutton and Barto(2018)}]{sutton2018reinforcement}
\bibinfo{author}{Sutton, R.S.}, \bibinfo{author}{Barto, A.G.},
  \bibinfo{year}{2018}.
\newblock \bibinfo{title}{Reinforcement learning: An introduction}.
\newblock \bibinfo{publisher}{MIT press}, \bibinfo{address}{Cambridge, MA,
  USA}.
\bibitem[{Swaney and Grossmann(1985)}]{swaney1985index}
\bibinfo{author}{Swaney, R.E.}, \bibinfo{author}{Grossmann, I.E.},
  \bibinfo{year}{1985}.
\newblock \bibinfo{title}{An index for operational flexibility in chemical
  process design. part i: Formulation and theory}.
\newblock \bibinfo{journal}{AIChE Journal} \bibinfo{volume}{31},
  \bibinfo{pages}{621--630}.
\bibitem[{Tawarmalani and Sahinidis(2005)}]{tawarmalani2005polyhedral}
\bibinfo{author}{Tawarmalani, M.}, \bibinfo{author}{Sahinidis, N.V.},
  \bibinfo{year}{2005}.
\newblock \bibinfo{title}{A polyhedral branch-and-cut approach to global
  optimization}.
\newblock \bibinfo{journal}{Mathematical programming} \bibinfo{volume}{103},
  \bibinfo{pages}{225--249}.
\bibitem[{Taylor and Stone(2007)}]{taylor2007cross}
\bibinfo{author}{Taylor, M.E.}, \bibinfo{author}{Stone, P.},
  \bibinfo{year}{2007}.
\newblock \bibinfo{title}{Cross-domain transfer for reinforcement learning},
  in: \bibinfo{booktitle}{Proceedings of the 24th International Conference on
  Machine Learning}, pp. \bibinfo{pages}{879--886}.
\bibitem[{Tewari et~al.(2020)Tewari, Liu and Papageorgiou}]{Tewari2020}
\bibinfo{author}{Tewari, A.}, \bibinfo{author}{Liu, K.H.},
  \bibinfo{author}{Papageorgiou, D.}, \bibinfo{year}{2020}.
\newblock \bibinfo{title}{Information-theoretic sensor planning for large-scale
  production surveillance via deep reinforcement learning}.
\newblock \bibinfo{journal}{Computers \& Chemical Engineering}
  \bibinfo{volume}{141}.
\bibitem[{Thebelt et~al.(2021)Thebelt, Kronqvist, Mistry, Lee, Sudermann-Merx
  and Misener}]{thebelt2020entmoot}
\bibinfo{author}{Thebelt, A.}, \bibinfo{author}{Kronqvist, J.},
  \bibinfo{author}{Mistry, M.}, \bibinfo{author}{Lee, R.M.},
  \bibinfo{author}{Sudermann-Merx, N.}, \bibinfo{author}{Misener, R.},
  \bibinfo{year}{2021}.
\newblock \bibinfo{title}{Entmoot: A framework for optimization over ensemble
  tree models}.
\newblock \bibinfo{journal}{Computers \& Chemical Engineering}
  \bibinfo{volume}{151}, \bibinfo{pages}{107343}.
\bibitem[{Thebelt et~al.(2022)Thebelt, Tsay, Lee, Sudermann-Merx, Walz, Tranter
  and Misener}]{thebelt2022multi}
\bibinfo{author}{Thebelt, A.}, \bibinfo{author}{Tsay, C.},
  \bibinfo{author}{Lee, R.M.}, \bibinfo{author}{Sudermann-Merx, N.},
  \bibinfo{author}{Walz, D.}, \bibinfo{author}{Tranter, T.},
  \bibinfo{author}{Misener, R.}, \bibinfo{year}{2022}.
\newblock \bibinfo{title}{Multi-objective constrained optimization for energy
  applications via tree ensembles}.
\newblock \bibinfo{journal}{Applied Energy} \bibinfo{volume}{306},
  \bibinfo{pages}{118061}.
\bibitem[{Troup and Georgakis(2013)}]{troup2013process}
\bibinfo{author}{Troup, G.M.}, \bibinfo{author}{Georgakis, C.},
  \bibinfo{year}{2013}.
\newblock \bibinfo{title}{Process systems engineering tools in the
  pharmaceutical industry}.
\newblock \bibinfo{journal}{Computers \& Chemical Engineering}
  \bibinfo{volume}{51}, \bibinfo{pages}{157--171}.
\bibitem[{Tsay and Baldea(2019)}]{Tsay2019}
\bibinfo{author}{Tsay, C.}, \bibinfo{author}{Baldea, M.}, \bibinfo{year}{2019}.
\newblock \bibinfo{title}{110th anniversary: Using data to bridge the time and
  length scales of process systems}.
\newblock \bibinfo{journal}{Industrial \& Engineering Chemistry Research}
  \bibinfo{volume}{58}, \bibinfo{pages}{16696--16708}.
\bibitem[{Tsay and Baldea(2020)}]{tsay2020integrating}
\bibinfo{author}{Tsay, C.}, \bibinfo{author}{Baldea, M.}, \bibinfo{year}{2020}.
\newblock \bibinfo{title}{Integrating production scheduling and process control
  using latent variable dynamic models}.
\newblock \bibinfo{journal}{Control Engineering Practice} \bibinfo{volume}{94},
  \bibinfo{pages}{104201}.
\bibitem[{Tsay et~al.(2021)Tsay, Kronqvist, Thebelt and
  Misener}]{tsay2021partition}
\bibinfo{author}{Tsay, C.}, \bibinfo{author}{Kronqvist, J.},
  \bibinfo{author}{Thebelt, A.}, \bibinfo{author}{Misener, R.},
  \bibinfo{year}{2021}.
\newblock \bibinfo{title}{Partition-based formulations for mixed-integer
  optimization of trained {ReLU} neural networks}.
\newblock \bibinfo{journal}{Advances in Neural Information Processing Systems}
  \bibinfo{volume}{34}.
\bibitem[{Tsay et~al.(2019)Tsay, Kumar, Flores-Cerrillo and
  Baldea}]{tsay2019optimal}
\bibinfo{author}{Tsay, C.}, \bibinfo{author}{Kumar, A.},
  \bibinfo{author}{Flores-Cerrillo, J.}, \bibinfo{author}{Baldea, M.},
  \bibinfo{year}{2019}.
\newblock \bibinfo{title}{Optimal demand response scheduling of an industrial
  air separation unit using data-driven dynamic models}.
\newblock \bibinfo{journal}{Computers \& Chemical Engineering}
  \bibinfo{volume}{126}, \bibinfo{pages}{22--34}.
\bibitem[{Tsay et~al.(2018)Tsay, Pattison, Piana and Baldea}]{tsay2018survey}
\bibinfo{author}{Tsay, C.}, \bibinfo{author}{Pattison, R.C.},
  \bibinfo{author}{Piana, M.R.}, \bibinfo{author}{Baldea, M.},
  \bibinfo{year}{2018}.
\newblock \bibinfo{title}{A survey of optimal process design capabilities and
  practices in the chemical and petrochemical industries}.
\newblock \bibinfo{journal}{Computers \& Chemical Engineering}
  \bibinfo{volume}{112}, \bibinfo{pages}{180--189}.
\bibitem[{do~Valle et~al.(2018)do~Valle, de~Ara{\'u}jo~Kalid, Secchi and
  Kiperstok}]{do2018collection}
\bibinfo{author}{do~Valle, E.C.}, \bibinfo{author}{de~Ara{\'u}jo~Kalid, R.},
  \bibinfo{author}{Secchi, A.R.}, \bibinfo{author}{Kiperstok, A.},
  \bibinfo{year}{2018}.
\newblock \bibinfo{title}{Collection of benchmark test problems for data
  reconciliation and gross error detection and identification}.
\newblock \bibinfo{journal}{Computers \& Chemical Engineering}
  \bibinfo{volume}{111}, \bibinfo{pages}{134--148}.
\bibitem[{Vaupel et~al.(2020)Vaupel, Hamacher, Caspari, Mhamdi, Kevrekidis and
  Mitsos}]{vaupel2020accelerating}
\bibinfo{author}{Vaupel, Y.}, \bibinfo{author}{Hamacher, N.C.},
  \bibinfo{author}{Caspari, A.}, \bibinfo{author}{Mhamdi, A.},
  \bibinfo{author}{Kevrekidis, I.G.}, \bibinfo{author}{Mitsos, A.},
  \bibinfo{year}{2020}.
\newblock \bibinfo{title}{Accelerating nonlinear model predictive control
  through machine learning}.
\newblock \bibinfo{journal}{Journal of Process Control} \bibinfo{volume}{92},
  \bibinfo{pages}{261--270}.
\bibitem[{Venkatasubramanian(2019)}]{Venkatasubramanian2019}
\bibinfo{author}{Venkatasubramanian, V.}, \bibinfo{year}{2019}.
\newblock \bibinfo{title}{The promise of artificial intelligence in chemical
  engineering: Is it here, finally?}
\newblock \bibinfo{journal}{AIChE Journal} \bibinfo{volume}{65},
  \bibinfo{pages}{466--478}.
\bibitem[{Venkatasubramanian and Chan(1989)}]{Venkatasubramanian1989}
\bibinfo{author}{Venkatasubramanian, V.}, \bibinfo{author}{Chan, K.},
  \bibinfo{year}{1989}.
\newblock \bibinfo{title}{A neural network methodology for process fault
  diagnosis}.
\newblock \bibinfo{journal}{AIChE Journal} \bibinfo{volume}{35},
  \bibinfo{pages}{1993--2002}.
\bibitem[{Venkatasubramanian et~al.(2003)Venkatasubramanian, Rengaswamy, Kavuri
  and Yin}]{venkatasubramanian2003review}
\bibinfo{author}{Venkatasubramanian, V.}, \bibinfo{author}{Rengaswamy, R.},
  \bibinfo{author}{Kavuri, S.N.}, \bibinfo{author}{Yin, K.},
  \bibinfo{year}{2003}.
\newblock \bibinfo{title}{A review of process fault detection and diagnosis:
  Part iii: Process history based methods}.
\newblock \bibinfo{journal}{Computers \& Chemical Engineering}
  \bibinfo{volume}{27}, \bibinfo{pages}{327--346}.
\bibitem[{Voelker et~al.(2013)Voelker, Kouramas and
  Pistikopoulos}]{voelker2013simultaneous}
\bibinfo{author}{Voelker, A.}, \bibinfo{author}{Kouramas, K.},
  \bibinfo{author}{Pistikopoulos, E.N.}, \bibinfo{year}{2013}.
\newblock \bibinfo{title}{Simultaneous design of explicit/multi-parametric
  constrained moving horizon estimation and robust model predictive control}.
\newblock \bibinfo{journal}{Computers \& Chemical Engineering}
  \bibinfo{volume}{54}, \bibinfo{pages}{24--33}.
\bibitem[{Von~Stosch et~al.(2014)Von~Stosch, Oliveira, Peres and
  de~Azevedo}]{von2014hybrid}
\bibinfo{author}{Von~Stosch, M.}, \bibinfo{author}{Oliveira, R.},
  \bibinfo{author}{Peres, J.}, \bibinfo{author}{de~Azevedo, S.F.},
  \bibinfo{year}{2014}.
\newblock \bibinfo{title}{Hybrid semi-parametric modeling in process systems
  engineering: Past, present and future}.
\newblock \bibinfo{journal}{Computers \& Chemical Engineering}
  \bibinfo{volume}{60}, \bibinfo{pages}{86--101}.
\bibitem[{Walczak and Massart(2001)}]{Walczak2001}
\bibinfo{author}{Walczak, B.}, \bibinfo{author}{Massart, D.},
  \bibinfo{year}{2001}.
\newblock \bibinfo{title}{Dealing with missing data: Part i}.
\newblock \bibinfo{journal}{Chemometrics and Intelligent Laboratory Systems}
  \bibinfo{volume}{58}, \bibinfo{pages}{15--27}.
\bibitem[{Wan et~al.(2005)Wan, Pekny and Reklaitis}]{wan2005simulation}
\bibinfo{author}{Wan, X.}, \bibinfo{author}{Pekny, J.F.},
  \bibinfo{author}{Reklaitis, G.V.}, \bibinfo{year}{2005}.
\newblock \bibinfo{title}{Simulation-based optimization with surrogate
  models—application to supply chain management}.
\newblock \bibinfo{journal}{Computers \& Chemical Engineering}
  \bibinfo{volume}{29}, \bibinfo{pages}{1317--1328}.
\bibitem[{Wang et~al.(2018)Wang, Wang, Zhang and Mao}]{Wang2018}
\bibinfo{author}{Wang, X.J.}, \bibinfo{author}{Wang, X.Y.},
  \bibinfo{author}{Zhang, Q.}, \bibinfo{author}{Mao, Z.Z.},
  \bibinfo{year}{2018}.
\newblock \bibinfo{title}{The soft sensor of the molten steel temperature using
  the modified maximum entropy based pruned bootstrap feature subsets ensemble
  method}.
\newblock \bibinfo{journal}{Chemical Engineering Science}
  \bibinfo{volume}{189}, \bibinfo{pages}{401--412}.
\bibitem[{Wang and Hong(2020)}]{wang2020reinforcement}
\bibinfo{author}{Wang, Z.}, \bibinfo{author}{Hong, T.}, \bibinfo{year}{2020}.
\newblock \bibinfo{title}{Reinforcement learning for building controls: The
  opportunities and challenges}.
\newblock \bibinfo{journal}{Applied Energy} \bibinfo{volume}{269},
  \bibinfo{pages}{115036}.
\bibitem[{Wang and Ierapetritou(2017)}]{wang2017novel}
\bibinfo{author}{Wang, Z.}, \bibinfo{author}{Ierapetritou, M.},
  \bibinfo{year}{2017}.
\newblock \bibinfo{title}{A novel feasibility analysis method for black-box
  processes using a radial basis function adaptive sampling approach}.
\newblock \bibinfo{journal}{AIChE Journal} \bibinfo{volume}{63},
  \bibinfo{pages}{532--550}.
\bibitem[{Wiebe et~al.(2020)Wiebe, Cec\'ilio, Dunlop and Misener}]{Wiebe2020}
\bibinfo{author}{Wiebe, J.}, \bibinfo{author}{Cec\'ilio, I.},
  \bibinfo{author}{Dunlop, J.}, \bibinfo{author}{Misener, R.},
  \bibinfo{year}{2020}.
\newblock \bibinfo{title}{A robust approach to warped {Gaussian}
  process-constrained optimization}.
\newblock \href{http://arxiv.org/abs/2006.08222}{{\tt arXiv:2006.08222}}.
\bibitem[{Wiebe et~al.(2018)Wiebe, Cecílio and Misener}]{Wiebe2018}
\bibinfo{author}{Wiebe, J.}, \bibinfo{author}{Cecílio, I.},
  \bibinfo{author}{Misener, R.}, \bibinfo{year}{2018}.
\newblock \bibinfo{title}{Data-driven optimization of processes with degrading
  equipment}.
\newblock \bibinfo{journal}{Industrial \& Engineering Chemistry Research}
  \bibinfo{volume}{57}, \bibinfo{pages}{17177--17191}.
\bibitem[{Wilson and Sahinidis(2017)}]{wilson2017alamo}
\bibinfo{author}{Wilson, Z.T.}, \bibinfo{author}{Sahinidis, N.V.},
  \bibinfo{year}{2017}.
\newblock \bibinfo{title}{The {ALAMO} approach to machine learning}.
\newblock \bibinfo{journal}{Computers \& Chemical Engineering}
  \bibinfo{volume}{106}, \bibinfo{pages}{785--795}.
\bibitem[{Xenos et~al.(2016a)Xenos, Kahrs, Cicciotti, Leira and
  Thornhill}]{xenos2016}
\bibinfo{author}{Xenos, D.P.}, \bibinfo{author}{Kahrs, O.},
  \bibinfo{author}{Cicciotti, M.}, \bibinfo{author}{Leira, F.M.},
  \bibinfo{author}{Thornhill, N.F.}, \bibinfo{year}{2016}a.
\newblock \bibinfo{title}{Challenges of the application of data-driven models
  for the real-time optimization of an industrial air separation plant}, in:
  \bibinfo{booktitle}{2016 European Control Conference (ECC)},
  \bibinfo{organization}{IEEE}. pp. \bibinfo{pages}{1025--1030}.
\bibitem[{Xenos et~al.(2016b)Xenos, Noor, Matloubi, Cicciotti, Haugen and
  Thornhill}]{xenos2016b}
\bibinfo{author}{Xenos, D.P.}, \bibinfo{author}{Noor, I.M.},
  \bibinfo{author}{Matloubi, M.}, \bibinfo{author}{Cicciotti, M.},
  \bibinfo{author}{Haugen, T.}, \bibinfo{author}{Thornhill, N.F.},
  \bibinfo{year}{2016}b.
\newblock \bibinfo{title}{Demand-side management and optimal operation of
  industrial electricity consumers: An example of an energy-intensive chemical
  plant}.
\newblock \bibinfo{journal}{Appl. Energy} \bibinfo{volume}{182},
  \bibinfo{pages}{418--433}.
\bibitem[{Xu et~al.(2015)Xu, Lu, Baldea, Edgar, Wojsznis, Blevins and
  Nixon}]{Xu2015}
\bibinfo{author}{Xu, S.}, \bibinfo{author}{Lu, B.}, \bibinfo{author}{Baldea,
  M.}, \bibinfo{author}{Edgar, T.F.}, \bibinfo{author}{Wojsznis, W.},
  \bibinfo{author}{Blevins, T.}, \bibinfo{author}{Nixon, M.},
  \bibinfo{year}{2015}.
\newblock \bibinfo{title}{Data cleaning in the process industries}.
\newblock \bibinfo{journal}{Reviews in Chemical Engineering}
  \bibinfo{volume}{31}, \bibinfo{pages}{453--490}.
\bibitem[{Yan et~al.(2020)Yan, Borhani and Clough}]{yan2020machine}
\bibinfo{author}{Yan, Y.}, \bibinfo{author}{Borhani, T.},
  \bibinfo{author}{Clough, P.}, \bibinfo{year}{2020}.
\newblock \bibinfo{title}{Machine learning applications in chemical
  engineering}.
\newblock \bibinfo{journal}{Machine Learning in Chemistry}
  \bibinfo{volume}{17}, \bibinfo{pages}{340}.
\bibitem[{Yildirim et~al.(2017)Yildirim, Gebraeel and Sun}]{Yildirim2017}
\bibinfo{author}{Yildirim, M.}, \bibinfo{author}{Gebraeel, N.Z.},
  \bibinfo{author}{Sun, X.A.}, \bibinfo{year}{2017}.
\newblock \bibinfo{title}{Integrated predictive analytics and optimization for
  opportunistic maintenance and operations in wind farms}.
\newblock \bibinfo{journal}{IEEE Transactions on Power Systems}
  \bibinfo{volume}{32}, \bibinfo{pages}{4319--4328}.
\bibitem[{Yin et~al.(2014)Yin, Ding, Xie and Luo}]{yin2014review}
\bibinfo{author}{Yin, S.}, \bibinfo{author}{Ding, S.X.}, \bibinfo{author}{Xie,
  X.}, \bibinfo{author}{Luo, H.}, \bibinfo{year}{2014}.
\newblock \bibinfo{title}{A review on basic data-driven approaches for
  industrial process monitoring}.
\newblock \bibinfo{journal}{IEEE Transactions on Industrial Electronics}
  \bibinfo{volume}{61}, \bibinfo{pages}{6418--6428}.
\bibitem[{Yoon et~al.(2018)Yoon, Jordon and Schaar}]{Yoon2018}
\bibinfo{author}{Yoon, J.}, \bibinfo{author}{Jordon, J.},
  \bibinfo{author}{Schaar, M.V.D.}, \bibinfo{year}{2018}.
\newblock \bibinfo{title}{Gain: Missing data imputation using generative
  adversarial nets}.
\bibitem[{Zavala et~al.(2008)Zavala, Laird and Biegler}]{zavala2008fast}
\bibinfo{author}{Zavala, V.M.}, \bibinfo{author}{Laird, C.D.},
  \bibinfo{author}{Biegler, L.T.}, \bibinfo{year}{2008}.
\newblock \bibinfo{title}{A fast moving horizon estimation algorithm based on
  nonlinear programming sensitivity}.
\newblock \bibinfo{journal}{Journal of Process Control} \bibinfo{volume}{18},
  \bibinfo{pages}{876--884}.
\bibitem[{Zhang et~al.(2016a)Zhang, Grossmann, Sundaramoorthy and
  Pinto}]{zhang2016b}
\bibinfo{author}{Zhang, Q.}, \bibinfo{author}{Grossmann, I.E.},
  \bibinfo{author}{Sundaramoorthy, A.}, \bibinfo{author}{Pinto, J.M.},
  \bibinfo{year}{2016}a.
\newblock \bibinfo{title}{Data-driven construction of convex region surrogate
  models}.
\newblock \bibinfo{journal}{Optim. Eng.} \bibinfo{volume}{17},
  \bibinfo{pages}{289--332}.
\bibitem[{Zhang et~al.(2016b)Zhang, Sundaramoorthy, Grossmann and
  Pinto}]{zhang2016}
\bibinfo{author}{Zhang, Q.}, \bibinfo{author}{Sundaramoorthy, A.},
  \bibinfo{author}{Grossmann, I.E.}, \bibinfo{author}{Pinto, J.M.},
  \bibinfo{year}{2016}b.
\newblock \bibinfo{title}{A discrete-time scheduling model for continuous
  power-intensive process networks with various power contracts}.
\newblock \bibinfo{journal}{Computers \& Chemical Engineering}
  \bibinfo{volume}{84}, \bibinfo{pages}{382--393}.
\bibitem[{Zhang(2015)}]{Zhang2015}
\bibinfo{author}{Zhang, Y.}, \bibinfo{year}{2015}.
\newblock \bibinfo{title}{Wiener and {Gamma} processes overview for degradation
  modelling and prognostic}.
\newblock \bibinfo{journal}{NorwegianUniversity of Science and Technology} .
\bibitem[{Zhang and Dong(2014)}]{Zhang2014}
\bibinfo{author}{Zhang, Z.}, \bibinfo{author}{Dong, F.}, \bibinfo{year}{2014}.
\newblock \bibinfo{title}{Fault detection and diagnosis for missing data
  systems with a three time-slice dynamic {Bayesian} network approach}.
\newblock \bibinfo{journal}{Chemometrics and Intelligent Laboratory Systems}
  \bibinfo{volume}{138}, \bibinfo{pages}{30--40}.
\bibitem[{Zhao and Huang(2018)}]{Zhao2018}
\bibinfo{author}{Zhao, C.}, \bibinfo{author}{Huang, B.}, \bibinfo{year}{2018}.
\newblock \bibinfo{title}{A full-condition monitoring method for nonstationary
  dynamic chemical processes with cointegration and slow feature analysis}.
\newblock \bibinfo{journal}{AIChE Journal} \bibinfo{volume}{64},
  \bibinfo{pages}{1662--1681}.
\bibitem[{Zhao and You(2020)}]{Zhao2020}
\bibinfo{author}{Zhao, S.}, \bibinfo{author}{You, F.}, \bibinfo{year}{2020}.
\newblock \bibinfo{title}{Distributionally robust chance constrained
  programming with generative adversarial networks ({GAN}s)}.
\newblock \bibinfo{journal}{AIChE Journal} \bibinfo{volume}{66}.
\bibitem[{Zhao et~al.(2014)Zhao, Huang and Liu}]{zhao2014constrained}
\bibinfo{author}{Zhao, Z.}, \bibinfo{author}{Huang, B.}, \bibinfo{author}{Liu,
  F.}, \bibinfo{year}{2014}.
\newblock \bibinfo{title}{Constrained particle filtering methods for state
  estimation of nonlinear process}.
\newblock \bibinfo{journal}{AIChE Journal} \bibinfo{volume}{60},
  \bibinfo{pages}{2072--2082}.
\bibitem[{Ławryńczuk(2008)}]{Lawrynczuk2008}
\bibinfo{author}{Ławryńczuk, M.}, \bibinfo{year}{2008}.
\newblock \bibinfo{title}{Modelling and nonlinear predictive control of a yeast
  fermentation biochemical reactor using neural networks}.
\newblock \bibinfo{journal}{Chemical Engineering Journal}
  \bibinfo{volume}{145}, \bibinfo{pages}{290--307}.

\end{thebibliography}

\end{document}